\pdfoutput=1
\documentclass[11pt]{article}

\usepackage{acl}
\usepackage{xspace}

\usepackage{times}
\usepackage{latexsym}
\usepackage{bm}
\usepackage{pifont}
\usepackage{float}       
\usepackage{enumitem}
\usepackage{subfig}                
\usepackage{overpic}               
\usepackage{pgfplots}
\usepackage{soul}

\usepackage[hyphenbreaks]{breakurl}

\usepackage[T1]{fontenc}
\usepackage[utf8]{inputenc}

\usepackage{microtype}

\usepackage{inconsolata}

\usepackage{amsmath}
\usepackage{amssymb}
\usepackage{booktabs}
\usepackage{tikz}
\usepackage{makecell}
\usepackage{graphicx}
\usepackage{multicol}
\usepackage{multirow}
\usepackage{subcaption}

\title{PartialFormer: Modeling Part Instead of Whole for Machine Translation}

\author{
Tong Zheng$^{1}$\thanks{\xspace\xspace Equal Contribution.
},
  Bei Li$^{1*}$,
  Huiwen Bao$^{1,2*}$,
  Jiale Wang$^{1}$,
  Weiqiao Shan$^{1}$,\\
  \textbf{Tong Xiao$^{1,2}$\thanks{\xspace\xspace Corresponding author.}~~}
  \textbf{and Jingbo Zhu$^{1,2}$}\\
  $^1$School of Computer Science and Engineering, Northeastern University, Shenyang, China\\
  $^2$NiuTrans Research, Shenyang, China \\
  {\tt
        \{zhengtong12356, goodbaohuiwen\}@gmail.com, libei\_neu@outlook.com
  }\\
  {\tt
        \{xiaotong,zhujingbo\}@mail.neu.edu.cn
  }
}

\begin{document}
\maketitle
\begin{abstract}
The design choices in Transformer feed-forward neural networks have resulted in significant computational and parameter overhead. In this work, we emphasize the importance of hidden dimensions in designing lightweight FFNs, a factor often overlooked in previous architectures. Guided by this principle, we introduce PartialFormer, a parameter-efficient Transformer architecture utilizing multiple smaller FFNs to reduce parameters and computation while maintaining essential hidden dimensions. These smaller FFNs are integrated into a multi-head attention mechanism for effective collaboration. We also propose a tailored head scaling strategy to enhance PartialFormer's capabilities. Furthermore, we present a residual-like attention calculation to improve depth scaling within PartialFormer. Extensive experiments on 9 translation tasks and 1 abstractive summarization task validate the effectiveness of our PartialFormer approach on machine translation and summarization tasks. 
Our code would be available
at: \url{https://github.com/zhengkid/PartialFormer}.
\end{abstract}

\section{Introduction}

The Transformer model~\cite{Vaswani2017transformer} has emerged as a cornerstone in the natural language processing~(NLP) domain, overshadowing convolutional neural networks~\cite{Gehring2017convolutionals2s} and recurrent neural networks~\cite{Sutskever2014sequence2sequence} by virtue of its minimal inductive bias, superior scalability, and proficiency in modeling sequences. Nonetheless, its substantial computational and parametric requisites pose significant challenges to its deployment and training, warranting an ongoing trend in the research community toward eliminating redundant parameters and computations~\cite{Dehghani2019UniversalTransformer,Mehta2019DeFINEDF, Lan2020ALBERT, Wu2020Lite, Mehta2021Delight,reid-etal-2021-subformer-exploring,li-etal-2022-ode} in Transformer.

\begin{figure}[t!]
	\centering
        \usetikzlibrary{shapes.geometric}
          \tikzset{global scale/.style={
    scale=#1,
    every node/.append style={scale=#1}
  }
}
        \begin{tikzpicture}[global scale=0.8]

        \def\nodehsep{1.2em}
        \def\wh{2em}
        \def\ww{1em}
        \def\whsep{0.4em}
        \def\sww{2em}
        \def\swwsep{2.5em}
        \def\reluw{5.5em}
        \def\reluh{1em}
        \def\verytiny{\fontsize{2pt}{5pt}\selectfont}
        
        % Vanilla FFN
        \begin{scope}[xshift=0em]
        {\small
        \node [anchor=north,trapezium, trapezium left angle=120, trapezium right angle=120, minimum width=\ww, minimum height=\wh,fill=orange!30] (w11) at (0, 0) {$W_1$};

        \node [anchor=south,font=\footnotesize,rectangle, minimum width=\reluw, minimum height=\reluh,fill=blue!20,inner sep=0pt](relu) at ([yshift=\nodehsep]w11.north) {ReLU};

        \node [anchor=south,trapezium, trapezium left angle=60, trapezium right angle=60, minimum width=\ww, minimum height=\wh,fill=orange!30] (w21) at ([yshift=\nodehsep]relu.north) {$W_2$};
        }

        {\scriptsize
        \node [anchor=north west, inner sep=0em] (s11) at ([xshift=0.5em,yshift=-\whsep]w11.south) {$d$};
        \node [anchor=south west, inner sep=0em] (s12) at ([xshift=0.5em,yshift=\whsep*0.7]w11.north) {$d_{\text{ffn}}$};
        \node [anchor=north west, inner sep=0em] (s21) at ([xshift=0.5em,yshift=-\whsep*0.7]w21.south) {$d_{\text{ffn}}$};
        \node [anchor=south west, inner sep=0em] (s22) at ([xshift=0.5em,yshift=\whsep]w21.north) {$d$};

        \draw[->] ([yshift=-\nodehsep*0.9]w11.south) -- ([yshift=-0em]w11.south);
        \draw[->] ([yshift=-0em]w11.north) -- ([yshift=-0em]relu.south);
        \draw[->] ([yshift=-0em]relu.north) -- ([yshift=-0em]w21.south);
        }

        \node [anchor=center,rectangle,rounded corners=5pt,draw,dotted,minimum width=\reluw*0.95, minimum height=\wh*4](back) at (relu.center) {};

        \node [anchor=north, font=\footnotesize, inner sep=0em] (l) at ([yshift=-4em]relu.south) {(a) Vanilla};
        \end{scope}

        % Lightweight FFN
        \begin{scope}[xshift=0.85in]
        {\small
        \node [anchor=north,trapezium, trapezium left angle=60, trapezium right angle=60, minimum width=\ww, minimum height=\wh,fill=orange!30] (w11) at (0, 0) {$W_1$};

        \node [anchor=south,font=\footnotesize,rectangle, minimum width=\ww, minimum height=\reluh,fill=blue!20,inner sep=2pt](relu) at ([yshift=\nodehsep]w11.north) {ReLU};

        \node [anchor=south,trapezium, trapezium left angle=120, trapezium right angle=120, minimum width=\ww, minimum height=\wh,fill=orange!30] (w21) at ([yshift=\nodehsep]relu.north) {$W_2$};
        }

        {\scriptsize
        \node [anchor=north west, inner sep=0em] (s11) at ([xshift=0.5em,yshift=-\whsep]w11.south) {$d$};
        \node [anchor=south west, inner sep=0em] (s12) at ([xshift=0.5em,yshift=\whsep*0.7]w11.north) {$d_{\text{ffn}}$};
        \node [anchor=north west, inner sep=0em] (s21) at ([xshift=0.5em,yshift=-\whsep*0.7]w21.south) {$d_{\text{ffn}}$};
        \node [anchor=south west, inner sep=0em] (s22) at ([xshift=0.5em,yshift=\whsep]w21.north) {$d$};

        \draw[->] ([yshift=-\nodehsep*0.9]w11.south) -- ([yshift=-0em]w11.south);
        \draw[->] ([yshift=-0em]w11.north) -- ([yshift=-0em]relu.south);
        \draw[->] ([yshift=-0em]relu.north) -- ([yshift=-0em]w21.south);
        }

        \node [anchor=center,rectangle,rounded corners=5pt,draw,dotted,minimum width=\reluw*0.95, minimum height=\wh*4](back) at (relu.center) {};

        \node [anchor=north, font=\footnotesize, inner sep=0em] (l) at ([yshift=-4em]relu.south) {(b) Lightweight};
        \end{scope}

        % PartialFormer FFN
        \begin{scope}[xshift=1.53in]
        {\small
        \node [anchor=north,trapezium, trapezium angle=130, trapezium stretches body=true, minimum width=\sww, minimum height=\wh,fill=orange!30] (w11) at (0, 0) {};
        \node [anchor=center,trapezium, trapezium angle=130, trapezium stretches body=true, minimum width=\sww, minimum height=\wh,fill=orange!30] (w12) at ([xshift=\swwsep]w11.center) {};
        \node [anchor=west, font=\tiny, inner sep=0em,inner sep=0pt] (w13) at ([xshift=\swwsep*0.5]w12.center) {$\cdots$};
        \node [anchor=center,trapezium, trapezium angle=130, trapezium stretches body=true, minimum width=\sww, minimum height=\wh,fill=orange!30] (w14) at ([xshift=\swwsep*0.5]w13.east) {};

        \node [anchor=center, font=\small] (ww11) at (w11.center) {$W_1$};
        \node [anchor=center, font=\small] (ww12) at (w12.center) {$W_1$};
        \node [anchor=center, font=\small] (ww14) at (w14.center) {$W_1$};
        
        \node [anchor=south west,font=\footnotesize,rectangle, minimum width=\reluw*1.45, minimum height=\reluh,fill=blue!20,inner sep=0pt](relu) at ([xshift=-\sww*0.5,yshift=\nodehsep]w11.north) {ReLU};

        \node [anchor=south,trapezium, trapezium angle=60, trapezium stretches body=true, minimum width=\sww, minimum height=\wh,fill=orange!30] (w21) at ([xshift=\sww*0.5,yshift=\nodehsep]relu.north west) {};
        \node [anchor=center,trapezium, trapezium angle=60, trapezium stretches body=true, minimum width=\sww, minimum height=\wh,fill=orange!30] (w22) at ([xshift=\swwsep]w21.center) {};
        \node [anchor=west, inner sep=0em, font=\tiny] (w23) at ([xshift=\swwsep*0.5]w22.center) {$\cdots$};
        \node [anchor=center,trapezium, trapezium angle=60, trapezium stretches body=true, minimum width=\sww, minimum height=\wh,fill=orange!30] (w24) at ([xshift=\swwsep*0.5]w23.east) {};

        \node [anchor=center, font=\tiny] (ww21) at (w21.center) {$W_2$};
        \node [anchor=center, font=\tiny] (ww22) at (w22.center) {$W_2$};
        \node [anchor=center, font=\tiny] (ww24) at (w24.center) {$W_2$};
        }
        {\scriptsize
        \node [anchor=north west, inner sep=0em] (s11) at ([xshift=0.2em]w11.south) {$d^{'}$};
        \node [anchor=south west, inner sep=0em] (s12) at ([xshift=0.2em]w11.north) {$d^{'}_{\text{ffn}}$};
        \node [anchor=north west, inner sep=0em] (s21) at ([xshift=0.2em]w21.south) {$d^{'}_{\text{ffn}}$};
        \node [anchor=south west, inner sep=0em] (s22) at ([xshift=0.2em]w21.north) {$d^{'}$};

        \node [anchor=north west, inner sep=0em] (s11) at ([xshift=0.2em]w12.south) {$d^{'}$};
        \node [anchor=south west, inner sep=0em] (s12) at ([xshift=0.2em]w12.north) {$d^{'}_{\text{ffn}}$};
        \node [anchor=north west, inner sep=0em] (s21) at ([xshift=0.2em]w22.south) {$d^{'}_{\text{ffn}}$};
        \node [anchor=south west, inner sep=0em] (s22) at ([xshift=0.2em]w22.north) {$d^{'}$};

        \node [anchor=north west, inner sep=0em] (s11) at ([xshift=0.2em]w14.south) {$d^{'}$};
        \node [anchor=south west, inner sep=0em] (s12) at ([xshift=0.2em]w14.north) {$d^{'}_{\text{ffn}}$};
        \node [anchor=north west, inner sep=0em] (s21) at ([xshift=0.2em]w24.south) {$d^{'}_{\text{ffn}}$};
        \node [anchor=south west, inner sep=0em] (s22) at ([xshift=0.2em]w24.north) {$d^{'}$};

        \draw[->] ([yshift=-\nodehsep*0.9]w11.south) -- ([yshift=-0em]w11.south);
        \draw[->] ([yshift=-0em]w11.north) -- ([yshift=\nodehsep*1.2]w11.north);
        \draw[->] ([yshift=-\nodehsep*1.2]w21.south) -- ([yshift=-0em]w21.south);

        \draw[->] ([yshift=-\nodehsep*0.9]w12.south) -- ([yshift=-0em]w12.south);
        \draw[->] ([yshift=-0em]w12.north) -- ([yshift=\nodehsep*1.2]w12.north);
        \draw[->] ([yshift=-\nodehsep*1.2]w22.south) -- ([yshift=-0em]w22.south);

        \draw[->] ([yshift=-\nodehsep*0.9]w14.south) -- ([yshift=-0em]w14.south);
        \draw[->] ([yshift=-0em]w14.north) -- ([yshift=\nodehsep*1.2]w14.north);
        \draw[->] ([yshift=-\nodehsep*1.2]w24.south) -- ([yshift=-0em]w24.south);
        }

        \node [anchor=center,rectangle,rounded corners=5pt,draw,dotted,minimum width=\reluw*1.42, minimum height=\wh*4](back) at (relu.center) {};

        \node [anchor=north, font=\footnotesize, inner sep=0em] (l) at ([yshift=-4em]relu.south) {(c) PartialFormer};
        \end{scope}
       
        \end{tikzpicture}
	\caption{Illustration of our idea.}
 \label{fig:merit}
 \vspace{-1em}
\end{figure}
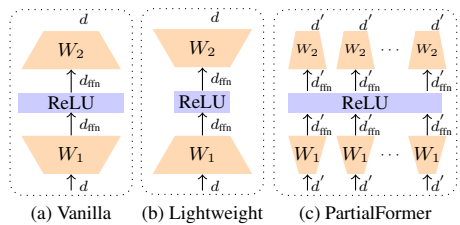

While these attempts represent significant strides in enhancing the efficiency of the Transformer architecture, they largely neglect an equally critical component: the Feed-Forward Network (FFN) that constitutes a substantial part of the Transformer's computational and parametric footprint, due to the inherent large feature space and hidden dimension.
Previous studies~\cite{Mehta2021Delight, Wu2020Lite, ge-etal-2022-edgeformer} have simplified FFNs by naively reducing their hidden dimensions, often at the expense of expressive power.
This leads to a question: \textit{Is the current formulation of lightweight FFNs truly optimal?}

To answer this concern, we turn to the insights provided by \citet{geva-etal-2021-transformer}, who depicted FFNs as a collection of key-value memories, where the number of memories is equal to the number of hidden dimensions in FFNs. This finding underscores the significance of hidden dimension in FFNs.
Drawing inspiration from this finding and the successful application of large hidden sizes in FFNs as evidenced by Meta's 4B model~\cite{tran-etal-2021-facebook}\footnote{They 
have shown enlarging the hidden size of FFNs to 16384 delivers significant BLEU improvements.}, 
we hypothesize that an efficient lightweight FFN is not merely about parameter reduction. Rather, it should aim to maintain or even increase the hidden dimension while judiciously reducing the number of parameters involved.

The literature on animal cognition provides some clues for designing lightweight and expressive FFNs. 
Research on animals' behavior has shown that group animals such as insects, fish, and some birds can emerge with some incredible abilities to deal with some complex tasks, though each individual owns poor abilities~\cite{couzin2009collective, conradt2005consensus}. 
This concept resonates with the AI community's ``Swarm Intelligence'' paradigm~\cite{Bonabeau1999SwarmIntelligence}, which emphasizes the power of collective decision-making. This biological prior motivates us to integrate Swarm Intelligence principles into the FFN design process.

To this end, we propose PartialFormer, an innovative approach to Transformer architecture. 
At the heart of PartialFormer lies the novel concept of Partial-Level Gated Feed-Forward Networks (PG-FFN). Conceived as an ensemble of streamlined FFNs operating in concert, each PG-FFN produces lower-dimensional hidden features. Despite their reduced individual dimensions, the aggregated output of these PG-FFNs either matches or surpasses the hidden dimensions of traditional, larger FFNs, as empirically substantiated in Figure \ref{fig:merit}. Moreover, we further equipped PartialFormer with a head scaling strategy tailed for efficiently scaling, and a residual-like attention calculation for stable optimization. 
These techniques empower PartialFormer to efficiently utilize parameters within the same parameter budget.

Our main contributions are as follows:
\begin{itemize}[itemsep=0pt,topsep=0pt,parsep=0pt]
    \item We introduced PG-FFNs, a method that efficiently reduces parameters and computations, and integrated them into the PartialFormer architecture for high performance. Additionally, we introduced an attention calculation method for stable optimization.
    \item We investigated the scalability of PartialFormer and proposed a head scaling strategy tailored for PartialFormer to efficient scaling.
    \item Rigorous empirical tests across 9 machine translation tasks and 1 abstractive summarization task confirm the effectiveness and efficiency of PartialFormer on machine translation and summarization tasks.
\end{itemize}

\section{Preliminary: Transformer}

In this section, we present some prior knowledge about the Transformer.
The Transformer block consists of a multi-head self-attention and a feed-forward network. Let $X\in\mathbb{R}^{T \times d}$ be a $T \times d$ input matrix of $T$ tokens. Each multi-head self-attention component owns $H$ heads. For simplicity, we omit layer normalization and residual connections.

\paragraph{Multi-Head Self-Attention} MHSA aims to model the global dependency among tokens. MHSA computes as follows:
\begin{eqnarray}
A_{i} &=& \mathrm{Softmax}(\frac{Q_{i}(K_i)^{\mathsf{T}}}{\sqrt{d_{k}}}), \label{eq:score_calculation}\\
\mathrm{head}_{i} &=& A_{i}V_{i}, \label{eq:agg} \\
X &=& \sum_{i=1}^{H} \mathrm{head}_{i}W^{O}_{i},  \label{eq:fusion}
 \label{eq:san}
\end{eqnarray}
\noindent
where $Q_i, K_i, V_i$ denote the query, key and value of $i$-th head, which are derived from input with three learnable matrices $W^Q_{i}, W^K_{i}, W^V_{i} \in \mathbb{R}^{d \times d_{k}}$ as follows: $Q_i=XW^Q_{i},K_i=XW^K_{i},V_i=XW^V_{i}$,  respectively. $W^{O}_{i} \in \mathbb{R}^{d_k \times d}$ is a learnable matrix. $d_k$ and $d$ denote the head dimension and embedding dimension, respectively. $A_{i}$ and $\mathrm{head}_{i}$  denote the attention matrix and representation of $i$-th head, respectively. 

\paragraph{Feed-Forward Network}  Feed-forward network is responsible for improving the expressiveness of the whole representation space by adopting an "expansion-activation-reduction" mapping strategy. It computes as follows:
\begin{equation}
X = \mathrm{ReLU}(XW_1 + b_1)W_2 + b_2,
\label{eq:ffn}
\end{equation}
\noindent where $W_1 \in \mathbb{R}^{d \times d_{\mathrm{ffn}}}, W_2 \in \mathbb{R}^{d_{\mathrm{ffn}} \times d}, b_1 \in \mathbb{R}^{d_{\mathrm{ffn}}}, b_2 \in \mathbb{R}^{d}$ are learnable matrices and $d_{\mathrm{ffn}}$ denotes the hidden dimension in FFN that is usually set to 4$d$.

\input{Figure/architecture_acl_version}

\section{PartialFormer}
\subsection{Overall Architecture}
Figure \ref{fig:architecture} illustrates the overall architecture of PartialFormer, encompassing both an encoder and a decoder. Although the foundational structure adheres to the design of the vanilla Transformer~\cite{Vaswani2017transformer}, there are some notable modifications.

\paragraph{Encoder.}
Different from vanilla Transformer, each encoder layer in PartialFormer consists of a unified sub-layer that integrates the PG-FFNs into the multi-head self-attention mechanism.

\paragraph{Decoder.}
Each decoder layer is composed of two types of sub-layers, both of which integrate the multi-head attention mechanism with PG-FFNs. The sub-layers differ based on the type of multi-head attention mechanisms employed, specifically whether it's a decoder self-attention or an encoder-decoder cross-attention mechanism. 
Notably, this design is inspired by previous studies~\cite{Lu-etal-2020-Macaron, Gulati-etal-2020-conformer}, but it differs in that we employ small FFNs, known as PG-FFNs, within each attention head of both the self-attention and cross-attention modules. To reduce computation, we halved the hidden dimension of PG-FFNs. Further decoder comparisons are in Appendix \ref{sec_compare_decoder}.

\subsection{Partial-Level Gated FFN}

\paragraph{Intuition}
Previous studies~\cite{Wu2020Lite, Mehta2021Delight, ge-etal-2022-edgeformer} commonly reduced the parameters in feed-forward networks by decreasing the hidden dimension (e.g., 2048 to 256). Different from them, 
our key idea involves utilizing a collection of small FFNs to model smaller input features expecting them to collaboratively emerge better performance while consuming fewer parameters, akin to ``Swarm Intelligence''.

In the concept of ``Swarm Intelligence'', a vanilla FFN can be viewed as a single large individual, which processes the whole feature input, making it, while effective, very resource-intensive in terms of computing power and memory. Assume a vanilla FFN with mappings of 1024->4096->1024, which consumes around 8.4 million parameters. By contrast, if we utilize multiple smaller FFNs (viewed as multiple weak individuals), each of which processes a subset of the input feature and collaboratively utilizes these outputs to generate the final output, the parameter and computation consumption will be significantly fewer. For example, 8 smaller FFNs with mappings of 128->512->128, we can retain the same hidden dimension, such as 8 * 512, while using only 1.05 million parameters. This approach significantly reduces parameters while maintaining the crucial hidden dimension, as emphasized in \citet{geva-etal-2021-transformer, tran-etal-2021-facebook}.

\paragraph{Design of PG-FFNs}
We have observed that the Transformer architecture inherently consists of multiple smaller subspaces, namely ``heads'' within the multi-head attention (MHA) mechanism. These heads act as sub-components of the original inputs and retain substantial information from the original data. Besides, the fusion mechanism in MHA enables the consolidation of the capabilities of multiple FFNs. As a result, PG-FFNs should naturally be constructed based on the MHA mechanism. More specifically, we insert multiple FFNs into the place between Eq. (\ref{eq:agg}) and Eq. (\ref{eq:fusion}), as shown in the blue part of Figure \ref{fig:architecture}(c).

While group transformation operations could be used to instantiate our idea, they are not optimal on GPUs due to their low I/O efficiency~\cite{Ningning2018shufflenetv2}, causing significant inference latency. To address this, we propose sharing parameters across each FFN within different heads, thereby eliminating the need for group transformation operations. However, directly sharing weights may result in homogeneous representations across different heads, which may potentially hinder the performance~\cite{li-etal-2018-multi-head}.  To mitigate this, we further introduce a head-specific gated mechanism. The core idea is to use a set of diverse masks to filter the information of different heads so that the head representation will be more diverse. 

Formally, given a set of head features $\{\mathrm{head}_{i}| 1 \leq i \leq H\}$ and diverse masks $\{G_{i}| 1 \leq i \leq H\}$, the calculation of PG-FFNs is as: 
\noindent
\begin{eqnarray}
\mathrm{\overline{head}}_{i}
&=& G_i\odot \mathrm{FFN}(\mathrm{head}_i),
 \label{eq:san}
\end{eqnarray}
\noindent
\noindent where $\mathrm{FFN}(\cdot)$  is the same as Eq. (\ref{eq:ffn}) and $G_i$ is generated via multiplication between the input feature of the block $X$ and a learnable matrix $W_i^G$ followed by an activation function $\sigma(\cdot)$, e.g., ReLU, Sigmoid and Tanh, as follows: $G_i = \sigma(XW_i^G)$. We compared the choice of $\sigma(\cdot)$ in Table \ref{tab:layer_ablation_en_de}.

\subsection{Residual-like Attention Calculation}
\citet{Dong2021PureAttention, Wang2022AntiOversmoothing} have shown that the original location of FFNs plays an essential role in optimizing transformers, e.g., alleviating \textit{Token Uniformity}. Therefore, it's vital to consider the impact of altering the FFN placement. Densely residual connections are effective but typically implemented either at the feature level (e.g., DLCL~\cite{wang-etal-2019-learning-deep}) or integrated into the network structure (e.g., Realformer~\cite{he-etal-2021-realformer}), which are not flexible.

To this end, we design a new variant of the residual connection integrated into the attention calculation, while also decoupling from the network architecture. Specifically, the calculation of attention maps consists of two parts: 1) $A^G$, the global part, and 2) $A^L$, the local part, as shown in Figure \ref{fig:architecture}(c). The calculation of $A^L$ remains the same as in the vanilla Transformer, while $A^G$ is computed once by using the original embedding as input through Eq. (\ref{eq:score_calculation}) (without softmax operation).
Inspired by \citet{he-etal-2021-realformer}, to efficiently fuse these components, we add them together and apply a Softmax function, as follows:
\begin{equation}
    A_i = \mathrm{Softmax}(A^G_{i} + A^L_i), \label{eq_residual_attention_calculation}
\end{equation}
where $A^G_i$ and $A^L_i$ denote the global and local attention map of $i$-th head.

In addition to the benefit of efficient depth scaling~(See Appendix \ref{sec:analysis_tu}), this approach provides remarkable flexibility in combining different attention mechanisms, specifically tailored to address specific conditions. For instance, it allows for the utilization of local attention to calculate $A^G$ when dealing with small datasets~(see Appendix \ref{sec:partial_different_A_G}).

\subsection{Efficient Scaling Strategy}
\label{subsec:scale_mechanism}
Though PG-FFN offers the advantage of reducing lots of parameters when applied directly to the transformer, it also leads to marginal performance degradation~(see Table \ref{tab:PartialFormer_VARIANTS} (a)). Thus, a crucial aspect of this study is to determine how to effectively utilize the spared parameters. In this work, we adopt a hybrid scaling strategy, which has been validated in computer vision, e.g., EfficientNet~\cite{tan2019efficientnet}. Note that our approach differs from EfficientNet, as we incorporate a combination of head scaling and depth scaling into our method.

\begin{table*}[t!]
    \centering
    \renewcommand{\arraystretch}{1}
\centering
\small

\setlength{\tabcolsep}{1.5pt}
\resizebox{0.83\linewidth}{!}{\begin{tabular}{clrrrrrccc}
\toprule

\bf Type &\bf Model & $\bm{N}$-$\bm{M}$& $\bm{d}$ & $\bm{d_k}$  & $\bm{H}$ & \bf Param & \bf BLEU & \bf COMET-22 & \bf sBLEU  \\
\midrule
\multirow{4}{*}{\bf \makecell{Multi-Branch \\ Architecture }} & Weighted Transformer~\cite{Ahmed2017WeightedTransformer} & 6-6 & 1024 &-&- & 211M & 28.90 & - & -\\
&Multi-Unit Transformer~\cite{yan-etal-2020-multi} & 6-6 & - & -  &  - & 130M & 29.30 & - & - \\
&MAT~\cite{Fan2020MultibranchAT} & 6-6 & - & - & -  & 206M & 29.90 & - & - \\
&Multi-Path Transformer~\cite{lin-etal-2022-multi-path} & 6-6 & - & - & -  & 193M & 29.68 & - & - \\
\midrule
\multirow{2}{*}{\bf \makecell{Lightweight \\ Architecture}}&Evolved Transformer~\cite{So2019EvolvedTransformer} & - & - & -&-  &64M & 28.20 & - & -\\
&Delight~\cite{Mehta2021Delight} & - & 640 &- &- &54M & 28.00 & - & -\\
\midrule
\multirow{5}{*}{\bf \makecell{Weight Sharing}}&Universal Transformer~\cite{Dehghani2019UniversalTransformer} & - & 1024 & -&-  &65M & 28.90 & - & - \\
&SubFormer~\cite{reid-etal-2021-subformer-exploring} & - & - & - &- &63M & 28.50 & - & -\\
&SubFormer-big~\cite{reid-etal-2021-subformer-exploring} & - & - & -&- &197M & 29.30 & - & - \\
&ODE Transformer (RK4) \dag~\cite{li-etal-2022-ode} & 6-6 & 512 & - & -  & 62M & 28.88 & 83.47 & 27.8 \\
&ODE Transformer (RK2, Learn.)~\dag~\cite{li-etal-2022-ode} & 24-6 & 512 & - & -  & 118M & 29.73 & 83.94 & 28.6 \\
\midrule
\multirow{3}{*}{\bf \makecell{Other \\ Comparisons}} 
&RealFormer~\cite{he-etal-2021-realformer} & 18-18 & 512 & 64 & 8  & 151M & 29.35 & - & -\\
&DMAN~\dag~\cite{fan-etal-2021-mask} & 6-6 & 512 & 64 & 8  & 62M & 27.54 & 82.27 & 26.4 \\
&Mega-Softmax~\dag ~\cite{Ma2022mega} & 6-6 & 512 & - & 1  & 64M & 28.11 & 82.79 & 27.0\\
\midrule
\multirow{10}{*}{\bf Our System} 
% &Transformer & 6 & 512 & 64 & 8 &  & 62M & 27.43 & \\
% &Transformer & 24-6 & 360 & 45 & 8-8 &  & 62M & 28.00 & 26.9 \\
&Transformer & 24-6 & 512 & 64 & 8-8 & 118M & 29.05 & 83.60 & 27.9 \\
&PartialFormer (w/o Head Scaling) &24-6 & 512 & 64 & 8-8 & 66M &  28.86 & 83.35 & 27.7
\\
&PartialFormer &24-6 & 512 & 64 & 24-16 & 115M &  30.09 & 84.17 & 29.0\\
\cmidrule(r){2-10}
&Transformer & 6-6 & 512 & 64 & 8-8 & 62M & 27.43 & 82.19 & 26.4 \\
& PartialFormer (w/o Head Scaling) &	6-6	& 512	& 64	& 8-8	& 42M &	27.15	& 81.75	& 26.1 \\
& PartialFormer &	6-6	& 512	& 64	& 24-16 & 63M &	28.60 &	83.21 & 27.5 \\
\cmidrule(r){2-10}
&Transformer & 24-6 & 360 & 45 & 8-8  & 62M & 28.00 & 82.72 & 27.0 \\
&PartialFormer (w/o Head Scaling) &24-6 & 360 & 45 & 8-8 & 36M &  27.88& 82.49 & 26.8\\
&PartialFormer &24-6 & 360 & 45 & 24-16 & 61M &  29.23 & 83.74 & 28.1\\
&PartialFormer &24-6 & 360 & 45 & 30-16 & 68M &  29.56 & 83.94 & 28.4 \\

\bottomrule
\end{tabular}}
    \caption{Results on the WMT'14 En-De task. For a more fair comparison, we also re-implemented some state-of-the-arts models with same data and training strategy, as indicated by \dag.}
    \label{tab:result_main_en_de}
    \vspace{-1em}
\end{table*}

\paragraph{Head Scaling}

As aforementioned, PartialFormer is guided by ``swarm intelligence'' and operates with small subspaces. Expanding the number and size of these subspaces intuitively augments PartialFormer's capabilities. In response to this insight, we introduced a head-scaling strategy tailored specifically for PartialFormer, involving the direct addition of more heads and the expansion of their width, effectively bolstering its performance.

To achieve this objective, we decouple the relationship between the number of heads and the embedding size, specifically $d_k \times H \neq d$. This approach shares similarities with methods discussed in \citet{Bhojanapalli2020LowRankBI}. However, it differs in its two-step process, which draws inspiration from the inherent redundancy observed in attention maps as discussed in \citet{michel2019sixteen, clark-etal-2019-bert, voita-etal-2019-analyzing, nguyen2022improving, zheng-etal-2024-eit}. Given values for $d_k$, $d$, and $H$, we first create intermediate values for $Q$ and $K$, and then we expand the attention maps to the desired number of heads using a robust MLP network. In the case of $V$, we generate them directly. This approach allows for the inclusion of more heads in PartialFormer while maintaining the same parameter budget.

We demonstrate that this scaling strategy is naturally well-suited for PartialFormer (see Section \ref{sec:analysis_scaling_strategy_partialformer}). Furthermore, it can also be regarded as a variation of width scaling, offering two significant advantages: 1) enabling flexible imbalanced computation distribution in encoder-decoder architecture, and 2) preventing an excessive distribution of parameters in the embedding and output layers.

\section{Experimental Setups}
We assess PartialFormer's performance across both machine translation and abstractive summarization tasks\footnote{We tested PartialFormer's performance in language modeling, with results in the Appendix.}. More details are given in Appendix \ref{sec:detailed_settup}.

\paragraph{Dataset.} For the machine translation task, we selected 9 datasets involving WMT'14 English-German (En-De), WMT'14 English-French (En-Fr), WMT'16 English-Romanian (En-Ro), and six translation tasks from WMT'17 benchmark. We preprocessed the raw data following the standard strategy. For the abstractive summarization task, we utilized the widely-used CNN-DailyMail dataset. We followed the same preprocessing approach as described in ~\citet{ott-etal-2019-fairseq}. We applied joint byte pair encoding (BPE)~\cite{sennrich-etal-2016-neural} with sizes of 32K for all the tasks except the En-Ro task (20K), and CNN-DailyMail~(30K).

\paragraph{Training \& Evaluation.}

We trained models on GeForce RTX 3090 cards via Fairseq~\cite{ott-etal-2019-fairseq} toolkit primarily following the training strategy in \citet{wang-etal-2019-learning-deep}. For machine translation evaluation, we utilized \textit{multi-BLEU}~\cite{papineni-etal-2002-bleu}, COMET-22~\cite{rei-etal-2022-comet} and sacreBLEU~\cite{post-2018-call} scores. Following \citet{wang-etal-2019-learning-deep}, beam sizes were 4, 4, and 5 for En-De, En-Fr, and En-Ro tasks respectively. \textit{Length\_penalty} of 0.6, 0.8, and 1.3 were applied to En-De, En-Fr, and En-Ro tasks respectively. For the WMT'17 benchmark, beam size and \textit{Length\_penalty} were set to 4 and 1, respectively. We used an ensemble of last 10 checkpoints. For abstractive summarization, we set beam size, \textit{Length\_penalty}, minimum length and maximum length to 4, 2.0, 55 and 140, respectively. The evaluation metric was F1-Rouge~\cite{lin-2004-rouge}(Rouge-1,
Rouge-2 and Rouge-L).

\section{Experiments}
\subsection{Machine Translation}
Table \ref{tab:result_main_en_de} presents the results for the WMT'14 En-De task. $N-M$, $d$, $d_k$, $H$ and sBLEU denote encoder-decoder depths, embedding dimension, head dimension, number of heads and SacreBLEU, respectively. We made the following observations:
\begin{itemize}[itemsep=0pt,topsep=0pt,parsep=0pt]
    \item PartialFormer achieves BLEU scores of 28.60, 29.56, and 30.09 in three different configurations, surpassing the standard Transformer by 1.17 BLEU points, 1.56 BLEU points, and 1.04 BLEU points with a similar model capacity. These observations are further supported by COMET-22 and sacreBLEU scores.

    \item Without the head scaling strategy, PartialFormer performs slightly worse than the standard Transformer (27.15 vs. 27.43, 27.88 vs. 28.00, and 28.86 vs. 29.05) but is significantly more parameter-efficient (42M vs. 62M, 36M vs. 62M, 66M vs. 118M). This is due to our PG-FFN structure, which maintains high hidden dimensions while reducing parameter usage.
 
    \item PartialFormer surpasses other multi-branch Transformers and state-of-the-art weight-sharing methods like ODE Transformer~\cite{li-etal-2022-ode}, as well as strong baselines such as Mega~\cite{Ma2022mega}. Notably, ODE Transformer and Mega use extra relative position encoding and require more computational resources. Moreover, while Mega and DMAN train for up to 500K updates and 220 epochs, achieving BLEU scores of 29.01 and 29.10, our strategy involves only 50K updates, leading to their sub-optimal scores of 28.11 and 27.54 under similar conditions.
\end{itemize}
Tables \ref{tab:result_main_en_fr}, \ref{tab:result_main_en_ro}, and \ref{tab:result_main_wmt17} showcase results for the WMT'14 En-Fr, WMT'16 En-Ro, and WMT'17 benchmarks, respectively. Similar trends are observed in these tasks as in the En-De task.
\paragraph{MACs Comparison.} Table \ref{tab:Macs_comparison} displayed the multiplication-addition operations~(MACs), a metric for measuring neural network computations, on the En-De task. We made the following observations: 1) A deeper and narrower Transformer architecture consumes fewer computations while exhibiting superior performance (\#1 vs. \#2), 2) PartialFormer achieves comparable performance to the vanilla Transformer with the same width and depth, while utilizing fewer computations and parameters (\#2 vs. \#3), and 3) Head scaling is an efficient scaling strategy for PartialFormer to significantly improve its capacity (1.68 BLEU points) by adding 1.7B MACs and 32M parameters~(\#3 vs. \#4).

\begin{table}[t!]
    \centering
    \renewcommand{\arraystretch}{1}
\centering
\small

\setlength{\tabcolsep}{1pt}
\resizebox{1\linewidth}{!}{\begin{tabular}{lrrrrrcc}
\toprule

\bf Model & $\bm{N}$& $\bm{d}$ & $\bm{d_k}$  & $\bm{H}$& \bf Param & \bf BLEU & \bf COMET-22  \\
\midrule
 Weighted Transformer~\citeyearpar{Ahmed2017WeightedTransformer} & 6 & - &-& - & 211M & 41.40 & - \\
Evolved Transformer~\citeyearpar{So2019EvolvedTransformer} & - & - &- &- &64M & 40.60 &- \\
Delight~\citeyearpar{Mehta2021Delight} & - & 640 & -&-  &54M & 40.50 & - \\
ODE Transformer~(RK4)~\citeyearpar{li-etal-2022-ode} & 6 & -  & - &  - & 69M & 42.56 & - \\
ODE Transformer~(RK2, Learn.)~\citeyearpar{li-etal-2022-ode} & 24  & - & - &  - & 123M & 43.48 & - \\
 Multi-Path Transformer~\citeyearpar{lin-etal-2022-multi-path} & - & - & - &  - & 168M & 42.44 & - \\
\midrule
Transformer & 24 & 512 & 64 & 8-8 & 120M & 42.33 & 85.62 \\
PartialFormer &24 & 512 & 64 & 24-18  & 119M &  43.10 & 86.34 \\
PartialFormer &24 & 512 & 64 & 24-24  & 127M &  43.29 & 86.61  \\
\midrule
Transformer & 6 & 512 & 64 & 8-8  & 63M & 40.79 & 84.27 \\
Transformer & 24 & 360 & 45 & 8-8  & 64M & 40.96 & 84.42\\
PartialFormer &24 & 360 & 45 & 24-18  & 63M &  42.16 & 85.61  \\
PartialFormer &24 & 360 & 45 & 24-24  & 67M &  42.39 & 85.74 \\

\bottomrule
\end{tabular}}
    \caption{Results on the WMT'14 En-Fr task.}
    \label{tab:result_main_en_fr}
\end{table}

\begin{table}[t!]
    \centering
    \renewcommand{\arraystretch}{1}
\centering
\small

\setlength{\tabcolsep}{1pt}
\resizebox{1\linewidth}{!}{\begin{tabular}{lrrrrrrcc}
\toprule

\bf Model & $\bm{N}$& $\bm{d}$ & $\bm{d_k}$  & $\bm{H}$ & \bf Param & \bf BLEU & \bf COMET-22   \\
\midrule
Delight~\cite{Mehta2021Delight} & - & 640 & -&-  &53M & 34.70 & - \\
Subformer~\cite{reid-etal-2021-subformer-exploring} & - & - & - & -  & 48M & 34.70 & -  \\
ODE Transformer~(RK2 $\gamma$)~\dag~\citeyearpar{li-etal-2022-ode} & 6 & 1024 & 64 & 16-16  & 192M & 35.00 & 82.63  \\
\midrule
Transformer & 24 & 512 & 64 & 8-8  &111M & 35.00 & 82.11 \\
% PartialFormer~(w/o Head Scaling) & 24 & 512 & 64 & 8-8  &59M & 35.07  \\
PartialFormer &24 & 320 & 40 & 24-24  & 48M &  35.30 & 82.52 \\
% PartialFormer &24 & 512 & 64 & 24-16  &  &    \\
\bottomrule
\end{tabular}}
    \caption{Results on the WMT'16 En-Ro task. \dag  denotes re-implementation with same data and training strategy. }
    \label{tab:result_main_en_ro}
\end{table}

\begin{table}[t!]
    \centering
    \renewcommand{\arraystretch}{1}
\centering
\small

\setlength{\tabcolsep}{0.8pt}
\resizebox{1\linewidth}{!}{\begin{tabular}{lccccccccc}
\toprule
 \multirow{2}{*}{\bf Model} & \multicolumn{2}{c}{\bf Fi$\longleftrightarrow$En} &  \multicolumn{2}{c}{\bf De$\longleftrightarrow$En}  & \multicolumn{2}{c}{\bf Lv$\longleftrightarrow$En} &  \multirow{2}{*}{\bf Avg.}   \\
 \cmidrule(r){2-3}  \cmidrule(r){4-5}  \cmidrule(r){6-7}  
 & \bf Fi$\rightarrow$ En & \bf En$\rightarrow$ Fi & \bf De$\rightarrow$ En & \bf En$\rightarrow$ De & \bf Lv$\rightarrow$ En & \bf En$\rightarrow$ Lv  \\
\midrule

Transformer  & 26.07  & 22.14 & 35.04 & 28.59  & 17.59 & 16.23  & 24.27 \\
% Attention Calibration~\cite{} & - & 26.67  & 22.55 & - & - &  &  & 18.71 & 16.83 \\
% PartialFormer~(w/o Head Scaling) & -& - & -  & 34.40 &  - &  - & & \bf - &  \bf - \\
PartialFormer & \bf 27.48& \bf 23.35  & \bf 35.60 &  \bf 29.91 & \bf 19.65 &  \bf 17.37 & \bf 25.56\\

\bottomrule
\end{tabular}}
    \caption{Results on the WMT'17 benchmark. PartialFormer has the same depth and $d$ as the Transformer but consumes 1M fewer parameters on average.}
    \label{tab:result_main_wmt17}
\end{table}

\subsection{Abstractive Summarization}
Table \ref{tab:result_as} exhibited results on the CNN-DailyMail task. We can see that PartialFormer achieves better performance, as evidenced by higher Rough-1, Rough-2, and Rough-L scores, despite having fewer parameters (37M vs. 61M). This highlights the efficiency and effectiveness of the PartialFormer architecture in this task.

\section{Analysis}

\begin{table}[t!]
    \centering
    \renewcommand{\arraystretch}{1}
\centering
\small
\setlength{\tabcolsep}{1.5pt}
\resizebox{\linewidth}{!}{\begin{tabular}{clrrrrrrc}
\toprule
\textbf{\#}&\textbf{Model} & $\bm{N}$-$\bm{M}$ &	$\bm{d}$	&$\bm{d_k}$ &	$\bm{H}$  &\bf MACs & \bf Param  & \bf BLEU  \\
% \multicolumn{5}{c}{\bf \textit{Previous NMT System}} \\
\midrule
1&Transformer & 6-6 & 512 & 64 & 8-8 &  9.9B &62M  & 27.43 \\
\midrule
2&Transformer & 24-6 & 360 & 45 & 8-8 &   6.3B &62M  & 28.00 \\
3&PartialFormer~(w/o hs)& 	24-6	& 360	& 45	&8-8	& 5.2B &36M	& 27.88	 \\
4&PartialFormer& 	24-6	& 360	& 45	&30-16	& 6.9B &68M	& \bf 29.56	 \\

\bottomrule
\end{tabular}}
    \caption{MACs denote the multiplication-addition operations. We compute
them via 20 source and target tokens following \citet{Mehta2021Delight}.}
    \label{tab:Macs_comparison}
\end{table}

\begin{table}[t!]
    \centering
    \renewcommand{\arraystretch}{1}
\centering
\small
\setlength{\tabcolsep}{1.5pt}
\resizebox{\linewidth}{!}{\begin{tabular}{lrrrrrccc}
\toprule
\textbf{Model} & $\bm{N}$-$\bm{M}$ &	$\bm{d}$	&$\bm{d_k}$ &	$\bm{H}$  & \bf Param  & \bf RG-1 & \bf RG-2 & \bf RG-L \\
% \multicolumn{5}{c}{\bf \textit{Previous NMT System}} \\
\midrule
Transformer & 6-6 & 512 & 64 & 8-8 &61M  & 41.21 & 18.32 & 37.83\\
% PartialFormer& 	6-6	& 360	& 45	&30-16	&31M	& 41.19	& \bf 18.34	& \bf 37.93 \\
PartialFormer& 	6-6	& 400	& 50	&24-16	&37M	& \bf 41.50	& \bf 18.60	& \bf 38.25 \\

\bottomrule
\end{tabular}}
    \caption{Rough-1, Rough-2 and Rough-L comparisons on CNN-DailyMail task. }
    \label{tab:result_as}
\end{table}

\begin{table}[t!]
    \centering
    \renewcommand{\arraystretch}{1}
\centering
\small
\setlength{\tabcolsep}{2pt}

\resizebox{0.9\linewidth}{!}{\begin{tabular}{rlrc}
\toprule

\textbf{\#}&\textbf{Model}  & \bf Param & \bf BLEU   \\

% \multicolumn{5}{c}{\bf \textit{Previous NMT System}} \\
\midrule
1 &Transformer ($N=24$, $d=360$)  &62M  & 28.00 \\
2 &Pure Attention ($N=24$, $d=360$)  &31M  & 25.70 \\
\midrule
3 &PartialFormer  &68M  &\bf 29.56  \\
4 & \quad w/o Partial-level   Gated FFN  & 52M & \underline{27.51}\\
5 & \quad w/o Residual-like Attention Calculation  & 66M & 29.26\\
6 & \quad w/o Head Scaling  & 36M & 27.88\\
\midrule
7 & PartialFormer~(encoder only)  & 67M & 29.15\\
8 & PartialFormer~(decoder only)  & 63M & 28.80\\
\midrule
9 & PG-FFNs with Sigmoid activation &68M & 29.21 \\
10 & PG-FFNs with Tanh activation  &68M & 29.03  \\
\bottomrule
\end{tabular}}
    \caption{Ablation studies on WMT'14 En-De task.}
    \label{tab:layer_ablation_en_de}
\end{table}

\subsection{Ablation Studies}

Table \ref{tab:layer_ablation_en_de} presents an ablation study of PartialFormer on the WMT'14 En-De task, demonstrating the critical role of each component. Omitting any element causes performance decline, underscoring the holistic design. The PG-FFN removal (\#3 vs. \#4) results in a large performance drop of 2.05 BLEU points, despite a mere 16 million parameters reduction. This evidence corroborates previous findings~\cite{Dong2021PureAttention} on the subpar performance of pure attention networks sans FFN, highlighting the essential role of PG-FFN in PartialFormer.

Besides, Table \ref{tab:layer_ablation_en_de} shows the results of different PartialFormer configurations on the WMT'14 En-De task. The encoder-decoder PartialFormer achieves the highest performance, reaching 29.56 BLEU points, indicating the effectiveness of our approach in enhancing both the encoder and the decoder. Employing our concept to either the encoder or the decoder individually also improves performance, yet the encoder-decoder configuration persistently surpasses others, marking the greatest performance improvement.

\subsection{Comparison of Gating Strategy}

Table \ref{tab:layer_ablation_en_de}~(\#9 and \#10) presents a comparison of various activation functions used in PG-FFN. The results indicate that the default choice, ReLU activation, yields the best performance. 
One explanation is that the ReLU activation provides hard masks for filtering the information of different heads, compared to other activation functions. Such hard masks can make different heads more diverse.

\begin{table}[t!]
    \centering
    \renewcommand{\arraystretch}{1}
\centering
\small
\setlength{\tabcolsep}{5pt}
\resizebox{0.7\linewidth}{!}{
\begin{tabular}{cccccrc}
\toprule
\bf Setting & $\bm{H}$ & $\bm{d}$ & $\bm{d_k}$ & \bf Param & \bf BLEU \\

\midrule
\multirow{3}{*}{\makecell{Varying Encoder $H$}} &30-16& 360 & 45 &68M & 29.56 \\
&24-16& 360 & 45 &61M & 29.23 \\
&16-16& 360 & 45 &51M & 29.02 \\
\midrule
\multirow{3}{*}{\makecell{Varying Decoder $H$}} &16-16& 360 & 45 &51M & 29.02 \\
&16-24& 360 & 45 &56M & 28.85 \\
  &16-30& 360 & 45 & 60M & 29.20 \\
\midrule
  \multirow{3}{*}{Varying $d_k$}&30-16& 360 & 30 & 49M &  28.70 \\
  &30-16& 360 & 60 & 86M & 29.68 \\
  &30-16& 360 & 90 & 124M & 30.00 \\
\midrule
\multirow{3}{*}{\makecell{Varying $d$}}&30-16& 180 & 45 &35M & 27.61 \\
  &30-16& 270 & 45 &51M & 28.80 \\
 &30-16& 450 & 45 & 84M & 29.41 \\
\bottomrule
\end{tabular}}
\caption{Parameters analysis on WMT'14 En-De task.}
\label{tab:PartialFormer_parameter_analysis}
\end{table}

\subsection{Hyper-Parameter Analysis}
Since the proposed method relies on multiple parameters, we conducted additional experiments and analyses with different hyper-parameters, including the number of heads, head dimensions, and embedding dimensions, to further strengthen the robustness of our findings.
Table \ref{tab:PartialFormer_parameter_analysis} presented the results on the WMT'14 En-De task. We can observe that PartialFormer demonstrates strong performance across various choices of $H$, $d_k$, and $d$. This suggests that the superiority of PartialFormer arises from its efficient architecture design rather than hyper-parameter optimization.

\subsection{Analysis of Scaling Approaches for PartialFormer}
\label{sec:analysis_scaling_strategy_partialformer}

\begin{figure}[t!]
    \centering
     \tikzset{global scale/.style={
    scale=#1,
    every node/.append style={scale=#1}
  }
}
  \begin{tikzpicture}[]
  \pgfplotsset{set layers}
     \scriptsize{

    \begin{axis}[
      align=center,
	 at={(0,0)},
      ymajorgrids,
      xmajorgrids,
      grid style=dashed,
      width=0.22\textwidth,
      height=.20\textwidth,
      xlabel={\small \makecell{Params~(M)\\(a)}},
      ylabel={\small{BLEU}},
      ylabel style={yshift=-2em},xlabel style={yshift=1.0em},
      yticklabel style={/pgf/number format/precision=0,/pgf/number format/fixed zerofill},
      ymin=25,ymax=30, ytick={ 26, 27, 28, 29},
      xmin=13,xmax=87,xtick={20, 40, 60, 80},
      legend style={
at={(0,0)},
anchor=north east,at={(axis description cs:0,-0.1)}, fill=none, draw=none,yshift=-0.5em,xshift=0.5em,inner sep=0pt,legend plot pos=right,font={\small},cells={anchor=west}, legend columns = 1, ,column sep=0pt}
      ]

      \addplot[dotted, red, mark=star,mark size=.5pt,thick,mark options={solid, fill=red,draw=red,line width=3.25pt}] coordinates { (25, 25.61) (36, 27.88) (52, 28.15) (69, 28.88)
      };\label{Deep_scaling}\addlegendentry{\scalebox{0.6}{Depth Scaling}}
e
\addplot[dotted, orange,mark=square,mark size=.5pt,thick,mark options={solid, fill=orange,draw=orange,line width=1.25pt}] coordinates { (25, 25.51) (42, 27.15) (83, 28.39)
      };\label{width_scaling}\addlegendentry{\scalebox{0.6}{Witdth Scaling}}
       
      \addplot[dotted,blue,mark=diamond*,mark size=.5pt,thick,mark options={solid, fill=blue,draw=blue,line width=1.25pt}] coordinates {  (25, 25.51) (36, 27.88)
     (51, 29.02) (68, 29.56)

      };\label{Hybrid_scaling}\addlegendentry{\scalebox{0.6}{Hybrid Scaling}}

      \end{axis}

    \begin{axis}[
      align=center,
	 at={(0.22\textwidth,0)},
      ymajorgrids,
      xmajorgrids,
      grid style=dashed,
      width=0.22\textwidth,
      height=.20\textwidth,
      legend style={at={(0.23,0.08)}, anchor=south west},
      xlabel={\small \makecell{Params~(M)\\(b)}},
      ylabel={\small{BLEU}},
      ylabel style={yshift=-2em},xlabel style={yshift=1.0em},
      yticklabel style={/pgf/number format/precision=0,/pgf/number format/fixed zerofill},
      ymin=27.2,ymax=29.8, ytick={ 28, 29},
      xmin=30,xmax=90,xtick={20, 40, 60, 80},
      legend style={
at={(0,0)},
anchor=north east,at={(axis description cs:0,-0.1)}, fill=none, draw=none,yshift=-0.5em,xshift=0.5em,inner sep=0pt,legend plot pos=right,font={\small},cells={anchor=west}, legend columns = 1, ,column sep=0pt}
      ]

      \addplot[dotted, red, mark=star,mark size=.5pt,thick,mark options={solid, fill=red,draw=red,line width=3.25pt}] coordinates { (62, 28.0) (83, 28.51)  
      };\label{Deep_scaling}\addlegendentry{\scalebox{0.6}{Transformer}}

      \addplot[dotted, blue,mark=diamond*,mark size=.5pt,thick,mark options={solid, fill=blue,draw=blue,line width=1.25pt}] coordinates {  (36, 27.88)
      (68, 29.56)

      };\label{PartialFormer}\addlegendentry{\scalebox{0.6}{PartialFormer}}

      \end{axis}

     }

  \end{tikzpicture}
\vskip -0.1in
    \caption{(a) Scaling Up PartialFormer with Different Methods. (b) Scaling Transformer and PartialFormer with Head Scaling. }
    \label{fig:comparsion_scaling_methods}
    % \vspace{-1em}
\end{figure}
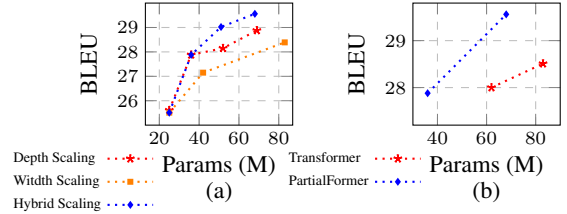

To disentangle the contribution of our proposed scaling
method from the PartialFormer architecture, Figure \ref{fig:comparsion_scaling_methods} compares the WMT'14 En-De performance of different scaling methods. Specifically, the initial setting is the PartialFormer ($N-M=6-6, H=8-8, d=360$). It's important to note that our hybrid scaling initially employs depth scaling, followed by head scaling.  
In general, all scaling methods improve BLEU scores with the cost of more parameters, but our hybrid scaling method can further improve BLEU, by up to 2.3\%, than other scaling methods, suggesting the importance of
our proposed hybrid scaling.

Head scaling can also improve the vanilla Transformer, though it is not as effective as in PartialFormer. Notably, PartialFormer attains 0.0525 BLEU per million parameters, significantly outperforming the vanilla Transformer (0.0243). This highlights the suitability of head-scaling for PartialFormer's design, a key contribution of this paper.

\definecolor{tiffanyblue}{RGB}{129,216,208}
\definecolor{bangdiblue}{RGB}{0,149,182}
\definecolor{kleinblue}{RGB}{0,47,167}
\definecolor{kabuliblue}{RGB}{26,85,153}
\definecolor{purple}{RGB}{138,43,226}

\subsection{Analysis on Behaviours of FFN}
\paragraph{Metric.} 

Following \citet{zhang-etal-2022-moefication}, we examine FFN behaviors across four aspects: activation neuron count (namely $n_{\mathrm{act.}}$), FFNs' hidden dimension, activation-neuron ratio (activations divided by hidden dimension, namely $R_{\mathrm{act.}}$), and FFN efficiency (activations divided by parameters, namely $\eta_{\mathrm{ffn}}$). Notably, for PartialFormer, the hidden dimension represents the concatenation of hidden dimensions from all smaller FFNs.

\paragraph{Results.}

\begin{figure*}[t!]
    \centering
  \begin{tikzpicture}[]
  \pgfplotsset{set layers}
     \scriptsize{

    \begin{axis}[
      align=center,
	 at={(0,0)},
      ymajorgrids,
      xmajorgrids,
      grid style=dashed,
      width=0.27\textwidth,
      height=.2\textwidth,
      legend style={at={(0.23,0.08)}, anchor=south west},
      xlabel={\small \makecell{Layer Index \\ (a)}},
      ylabel={\small{$n_{\mathrm{act.}}$}},
      ylabel style={yshift=-2em},xlabel style={yshift=1.0em},
      yticklabel style={/pgf/number format/precision=0,/pgf/number format/fixed zerofill},
      ymin=0,ymax=120, ytick={ 20, 40, 60, 80, 100},
      xmin=0,xmax=25,xtick={1, 6, 12, 18, 24},
      legend style={draw=none,yshift=6.5em,xshift=-3em,inner sep=0pt,legend plot pos=right,font={\small},cells={anchor=west}, legend columns = 2, ,column sep=2pt}
      ]

      \addplot[blue,mark=diamond*,mark size=0.25pt,thick,mark options={fill=blue,draw=blue,line width=1.25pt}] coordinates { (1, 50.397850036621094) (2, 46.05617141723633) (3, 30.716957092285156) (4, 32.27601623535156) (5, 28.592182159423828) (6, 25.675628662109375) (7, 24.729618072509766) (8, 28.869112014770508) (9, 26.933958053588867) (10, 30.077404022216797) (11, 33.576473236083984) (12, 31.970033645629883) (13, 32.3390998840332) (14, 33.447723388671875) (15, 33.69960021972656) (16, 31.835311889648438) (17, 31.542987823486328) (18, 34.97138595581055) (19, 35.815101623535156) (20, 36.96128845214844) (21, 40.59370803833008) (22, 43.48105239868164) (23, 50.58210754394531) (24, 62.86782455444336)

      };\label{baseline}
\addlegendentry{\scalebox{0.9}{Transformer~(BLEU: 28.00; Hidden Dimension: 1440)}}

      \addplot[red, mark=otimes*,mark size=0.25pt,thick,mark options={fill=red,draw=red,line width=1.25pt}] coordinates { (1, 29.44027328491211) (2, 34.75149917602539) (3, 13.85399341583252) (4, 30.174070358276367) (5, 101.7452621459961) (6, 19.441177368164062) (7, 4.418856620788574) (8, 49.83977508544922) (9, 32.83063888549805) (10, 15.632033348083496) (11, 40.52145767211914) (12, 46.91738510131836) (13, 89.28887176513672) (14, 74.86711883544922) (15, 58.14431381225586) (16, 53.80280685424805) (17, 58.88250732421875) (18, 51.280494689941406) (19, 55.328800201416016) (20, 42.71194076538086) (21, 57.96096420288086) (22, 38.90543746948242) (23, 66.41353607177734) (24, 41.248897552490234)
      };
      \addlegendentry{\scalebox{0.9}{PartialFormer~(BLEU: 29.56; Hidden Dimension: 5400)}}

\addplot[blue!30, line width=2pt] coordinates { (1, 35.7504) (24, 35.7504)
      };
      
        \addplot[red!30,line width=2pt] coordinates { (1, 46.1834)  (24, 46.1834)

      };

    % \draw[<-, thick] (axis cs:13, 46.1834) -- (axis cs:13,35.7504) ;
% \draw[->, thick] (axis cs:0, 0) -- (axis cs:5.5, 12.65) ;
% \node[anchor=west,font=\footnotesize] at (axis cs:13, 63.8563) {\bf 2.6$\mathbf{\times\uparrow}$ };

      \end{axis}

     \begin{axis}[
      align=center,
	 at={(0.25\textwidth,0)},
      ymajorgrids,
      xmajorgrids,
      grid style=dashed,
      width=0.27\textwidth,
      height=.2\textwidth,
      legend style={at={(0.23,0.08)}, anchor=south west},
      xlabel={\small \makecell{Layer Index \\ (b)}},
      ylabel={\small{$R_{\mathrm{act.}}$\tiny{(\%)}}},
      ylabel style={yshift=-3em},xlabel style={yshift=1.0em},
      yticklabel style={/pgf/number format/precision=0,/pgf/number format/fixed zerofill},
      ymin=0,ymax=5, ytick={ 1,2, 3,4},
      xmin=0,xmax=25,xtick={1, 6, 12, 18, 24},
      legend style={draw=none,yshift=8em,xshift=0em,inner sep=0pt,legend plot pos=right,font={\small},cells={anchor=west}, legend columns = 1, ,column sep=2pt}
      ]

      \addplot[blue,mark=diamond*,mark size=0.25pt,thick,mark options={fill=blue,draw=blue,line width=1.25pt}] coordinates { 
        (1, 3.628645658493042) (2, 3.3160417079925537) (3, 2.211622714996338) (4, 2.3238747119903564) (5, 2.058638334274292) (6, 1.8486480712890625) (7, 1.7805331945419312) (8, 2.0785739421844482) (9, 1.9392447471618652) (10, 2.1655755043029785) (11, 2.4175031185150146) (12, 2.3018369674682617) (13, 2.328415870666504) (14, 2.4082369804382324) (15, 2.4263756275177) (16, 2.292147159576416) (17, 2.2710978984832764) (18, 2.517946720123291) (19, 2.578690528869629) (20, 2.661208152770996) (21, 2.92274808883667) (22, 3.130636215209961) (23, 3.641909122467041) (24, 4.5264787673950195)

      };\label{baseline}
% \addlegendentry{\scalebox{1}{Transformer~(BLEU: 28.00)}}

      \addplot[red, mark=otimes*,mark size=0.25pt,thick,mark options={fill=red,draw=red,line width=1.25pt}] coordinates { (1, 0.5411818623542786) (2, 0.6388141512870789) (3, 0.2546690106391907) (4, 0.5546698570251465) (5, 1.870317816734314) (6, 0.3573746979236603) (7, 0.08122898638248444) (8, 0.916171669960022) (9, 0.6035030484199524) (10, 0.2873530089855194) (11, 0.744879961013794) (12, 0.8624525666236877) (13, 1.641342043876648) (14, 1.3762348890304565) (15, 1.068828821182251) (16, 0.9890232086181641) (17, 1.0823997259140015) (18, 0.9426538944244385) (19, 1.0170739889144897) (20, 0.78514564037323) (21, 1.065460205078125) (22, 0.7151732444763184) (23, 1.2208362817764282) (24, 0.7582511901855469)

      };
      % \addlegendentry{\scalebox{1}{PartialFormer~(BLEU: 29.58)}}

\addplot[blue!30, line width=2.5pt] coordinates { (1, 2.5740) (24, 2.5740)
      };
      
        \addplot[red!30,line width=2pt] coordinates { (1, 0.85)  (24, 0.85)

      };

    \draw[<-, thick] (axis cs:20, 0.85) -- (axis cs:20,2.5740) ;
% \draw[->, thick] (axis cs:0, 0) -- (axis cs:5.5, 12.65) ;
\node[anchor=west,font=\footnotesize] at (axis cs:9, 3.60) {\bf 3.0$\mathbf{\times\downarrow}$ };

      \end{axis}

          \begin{axis}[
      align=center,
	 at={(0.5\textwidth,0)},
      ymajorgrids,
      xmajorgrids,
      grid style=dashed,
      width=0.27\textwidth,
      height=.2\textwidth,
      legend style={at={(0.23,0.08)}, anchor=south west},
      xlabel={\small \makecell{Layer Index \\ (c)}},
      ylabel={\small{$\eta_{\mathrm{ffn}}$}\tiny{$(10^{-6})$}},
      ylabel style={yshift=-2em},xlabel style={yshift=1.0em},
      yticklabel style={/pgf/number format/precision=0,/pgf/number format/fixed zerofill},
      ymin=20,ymax=220, ytick={ 40, 80, 120, 160, 200},
      xmin=0,xmax=25,xtick={1, 6, 12, 18, 24},
      legend style={draw=none,yshift=8em,xshift=0em,inner sep=0pt,legend plot pos=right,font={\small},cells={anchor=west}, legend columns = 1, ,column sep=2pt}
      ]

      \addplot[blue,mark=diamond*,mark size=0.25pt,thick,mark options={fill=blue,draw=blue,line width=1.25pt}] coordinates { (1, 50.397850036621094) (2, 46.05617141723633) (3, 30.716957092285156) (4, 32.27601623535156) (5, 28.592182159423828) (6, 25.675628662109375) (7, 24.729618072509766) (8, 28.869112014770508) (9, 26.933958053588867) (10, 30.077404022216797) (11, 33.576473236083984) (12, 31.970033645629883) (13, 32.3390998840332) (14, 33.447723388671875) (15, 33.69960021972656) (16, 31.835311889648438) (17, 31.542987823486328) (18, 34.97138595581055) (19, 35.815101623535156) (20, 36.96128845214844) (21, 40.59370803833008) (22, 43.48105239868164) (23, 50.58210754394531) (24, 62.86782455444336)

      };\label{baseline}
% \addlegendentry{\scalebox{1}{Transformer~(BLEU: 28.00)}}

      \addplot[red, mark=otimes*,mark size=0.25pt,thick,mark options={fill=red,draw=red,line width=1.25pt}] coordinates { (1, 58.622589111328125) (2, 69.19847869873047) (3, 27.586591720581055) (4, 60.08373260498047) (5, 202.59915161132812) (6, 38.71195983886719) (7, 8.798989295959473) (8, 99.24288940429688) (9, 65.37349700927734) (10, 31.127098083496094) (11, 80.68778228759766) (12, 93.42373657226562) (13, 177.79580688476562) (14, 149.07830810546875) (15, 115.77934265136719) (16, 107.13433837890625) (17, 117.249267578125) (18, 102.11153411865234) (19, 110.17288970947266) (20, 85.04967498779297) (21, 115.41409301757812) (22, 77.47010803222656) (23, 132.24522399902344) (24, 82.13630676269531)

      };
      % \addlegendentry{\scalebox{1}{PartialFormer~(BLEU: 29.58)}}

\addplot[blue!30, line width=2.5pt] coordinates { (1, 35.7504) (24, 35.7504)
      };
      
        \addplot[red!30,line width=2.5pt] coordinates { (1, 91.9622)  (24, 91.9622)

      };

    \draw[<-, thick] (axis cs:13, 91.9622) -- (axis cs:13,35.7504) ;
% \draw[->, thick] (axis cs:0, 0) -- (axis cs:5.5, 12.65) ;
\node[anchor=west,font=\footnotesize] at (axis cs:13, 63.8563) {\bf 2.6$\mathbf{\times\uparrow}$ };

      \end{axis}

    \begin{axis}[
	 at={(0.75\textwidth,0)},
      ymajorgrids,
      xmajorgrids,
      grid style=dashed,
      width=0.27\textwidth,
      height=.2\textwidth,
      legend style={at={(0.23,0.08)}, anchor=south west},
      xlabel={\small \makecell{Layer Index
      \\ {(d) }}},
      ylabel={\small{$D_\mathrm{output}$}},
      ylabel style={yshift=-1.5em},xlabel style={yshift=1.0em},
      yticklabel style={/pgf/number format/precision=2,/pgf/number format/fixed zerofill},
      ymin=0.76,ymax=0.94, ytick={ 0.79, 0.82, 0.85, 0.88, 0.91},
      xmin=0,xmax=25,xtick={1, 6, 12, 18, 24},
      legend style={yshift=-0.5em,xshift=4em,inner sep=1pt,legend plot pos=right,font={\small},cells={anchor=west}}
      ]
      % \draw[|-|,line width=0.6pt, black!80, dashed, thick] (62,31.23) -- (110, 31.23);
      % using "mark options" do more changes for marks

      \addplot[blue,mark=diamond*,mark size=0.5pt,thick,mark options={fill=blue,draw=blue,line width=1.25pt}] coordinates { (1,0.7942273020744324)
(2,0.7962304353713989)
(3,0.7946145534515381)
(4,0.7945131659507751)
(5,0.7952606081962585)
(6,0.7940300703048706)
(7,0.7944667339324951)
(8,0.7958396077156067)
(9,0.7942259907722473)
(10,0.7952779531478882)
(11,0.7945087552070618)
(12,0.795943558216095)
(13,0.7943979501724243)
(14,0.7950928211212158)
(15,0.7946134805679321)
(16,0.7939436435699463)
(17,0.7947837114334106)
(18,0.7948220372200012)
(19,0.7949259281158447)
(20,0.7952647805213928)
(21,0.7957655787467957)
(22,0.7956266403198242)
(23,0.7946876287460327)
(24,0.7941218018531799)

      };

  \addplot[blue!30,line width=2.5pt] coordinates { (1,0.7949)
(24,0.7949)
      };

      \addplot[red, mark=otimes*,mark size=0.5pt,thick,mark options={fill=red,draw=red,line width=1.25pt}] coordinates { (1,0.9289955496788025)
(2,0.9303833246231079)
(3,0.9280907511711121)
(4,0.9178294539451599)
(5,0.9058080315589905)
(6,0.9134395718574524)
(7,0.9192654490470886)
(8,0.9071937799453735)
(9,0.9131340980529785)
(10,0.9085556268692017)
(11,0.9041228890419006)
(12,0.9006943106651306)
(13,0.906696081161499)
(14,0.9062924385070801)
(15,0.902535617351532)
(16,0.9111884832382202)
(17,0.9071165323257446)
(18,0.8982360363006592)
(19,0.9003955721855164)
(20,0.8912466764450073)
(21,0.8940209746360779)
(22,0.8901423811912537)
(23,0.894454300403595)
(24,0.8898761868476868)

      };

       \addplot[red!30, line width=2.5pt] coordinates { (1,0.9071)
(24,0.9071)
      };

      \draw[<-, thick] (axis cs:7, 0.9071) -- (axis cs:7,0.7949) ;
% \draw[->, thick] (axis cs:0, 0) -- (axis cs:5.5, 12.65) ;
\node[anchor=west,font=\footnotesize] at (axis cs:7, 0.8510) {\bf 1.1$\mathbf{\times \uparrow} $ };

      \end{axis}

     }

  \end{tikzpicture}
\vskip -0.1in
    \caption{Analysis on behaviours of FFNs and head diversity in Transformer and PartialFormer. }
    \label{fig:more_analysis}
    \vspace{-0.25em}
\end{figure*}
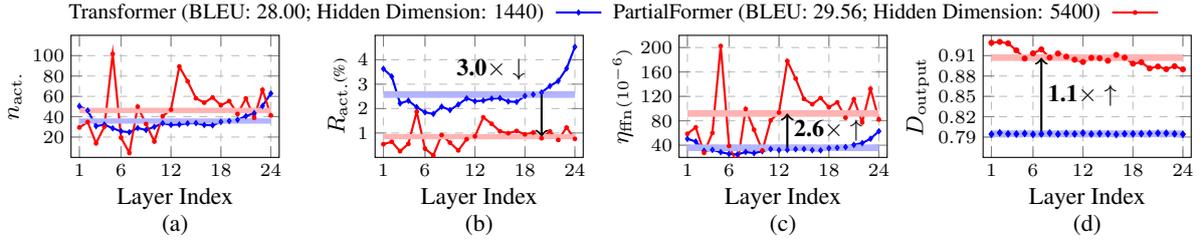

Figure \ref{fig:more_analysis} (a-c) exhibits the results on the En-De test set. It is evident that PartialFormer has a lower activation ratio than the vanilla Transformer, as shown in Figure \ref{fig:more_analysis} (b). This indicates that PG-FFNs present lower utilization of the hidden dimension compared to the vanilla FFNs. However, our PG-FFN is parameter consumption friendly, enabling larger hidden layer dimensions with the same parameter budget (e.g., 5400 vs. 1440). Despite lower utilization of hidden dimension, it can still own more activated neurons, as depicted in Figure \ref{fig:more_analysis} (a). Additionally, our PG-FFN exhibits higher efficiency compared to vanilla FFNs, as shown in Figure \ref{fig:more_analysis} (c).

\definecolor{tiffanyblue}{RGB}{129,216,208}
\definecolor{bangdiblue}{RGB}{0,149,182}
\definecolor{kleinblue}{RGB}{0,47,167}
\definecolor{kabuliblue}{RGB}{26,85,153}
\definecolor{purple}{RGB}{138,43,226}

\begin{table*}[t!]
\centering
\renewcommand{\arraystretch}{1}
\small
\setlength{\tabcolsep}{1.5pt}

\begin{minipage}[t]{0.5\textwidth}
\centering
\resizebox{\textwidth}{!}{\begin{tabular}{clcccc}
\toprule
\# & \bf Model & \bf \makecell{Batch Size}  & \bf Total Updates & \bf \makecell{Training Speed \\ (sec. / 100 updates)} & \bf BLEU \\
\midrule
1& Vanilla Transformer	&	8 x 4096 x 2	& 50K	& 28 & 28.00 \\
2 & Mega-Softmax & 8 x 8192 x 1 & 500K & 37& 29.01 \\
3 & Mega-Softmax (50k updates) 	& 8 x 8192 x 1	& 50K	&	37 &  28.11 \\
4 & ODE Transformer & 8 x 4096 x 2 & 50K & 34 & 29.03 \\
5 & ODE Transformer (reproduced) & 8 x 4096 x 2 & 50K & 34 & 28.88 \\
6 & SubFormer & 8 x 8192 x 2 & 250K (max) & 42 & 28.50 \\
7 & PartialFormer & 8 x 4096 x 2 & 50K & 40 &  29.56 \\
\bottomrule
\end{tabular}}
\caption*{(a) Training Phase}
\label{tab:result_convergence}
\end{minipage}%
\hspace{0.04\textwidth}
\begin{minipage}[t]{0.45\textwidth}
\centering
\resizebox{\textwidth}{!}{\begin{tabular}{llrccc}
\toprule
\#&\textbf{Model} &\bf Param 
& \bf Speed (Tok./s) & \bf Peak Memory& \bf COMET  \\
\midrule
\multicolumn{5}{c}{PartialFormer vs. vanilla Transformer}
\\
\midrule
1&Transformer&  62M &4325 &3.0G& 82.72 \\ 
2 &PartialFormer~(larger batch) & 36M & 6579&3.0G& 82.49 \\
\midrule
\multicolumn{5}{c}{PartialFormer vs. ODE Transformer} \\
\midrule
3&ODE Transformer&  118M &3254 &8.9G& 83.94 \\ 
4&PartialFormer & 68M &3023&3.3G& 83.94 \\
\bottomrule
\end{tabular}}
\caption*{(b) Inference Phase}
\label{tab:result_efficiency}
\end{minipage}

\caption{Efficiency analysis between PartialFormer and other Transformer variants.}
\label{tab:result_efficiency_analysis}
\end{table*}

\subsection{Analysis on Head Diversity}
\paragraph{Metric.} We select the same metric, namely $D_{output}$, as that in \citet{li-etal-2018-multi-head} to measure the diversity among head features. In this metric, a larger value indicates a higher level of diversity.
\paragraph{Results.}
From Figure \ref{fig:more_analysis} (d), we can observe that PartialFormer exhibits more diverse head features compared to the vanilla Transformer. This aligns with previous study~\cite{li-etal-2018-multi-head}, which demonstrates the positive impact of head feature diversity on the Transformer model's performance. Thus, we conclude that the insertion of FFNs into attention mechanism may be a more optimal design.

\subsection{Efficiency Analysis}
\paragraph{Convergence Analysis}
Table \ref{tab:result_efficiency_analysis} (a) compared the convergence updates, training speed and BLEU scores of PartialFormer with other methods. We made the following observations:
\begin{itemize}[itemsep=0pt,topsep=0pt,parsep=0pt]
    \item PartialFormer and ODE Transformer do not require more training updates to achieve higher performance than vanilla transformer, unlike other strong baselines. 
    \item All improved methods indeed lead to increased running latency. 
    \item PartialFormer achieve highest BLEU scores among all the comparisons.
\end{itemize}
Overall, we believe PartialFormer can achieve significant performance improvements while maintaining good training efficiency.
\paragraph{Inference Analysis}
Table \ref{tab:result_efficiency_analysis} (b) exhibits the inference efficiency on the test set of En-De task. We can see following observations: 1) Under the constraints of desired memory and performance, PartialFormer exhibits higher inference efficiency (6579 vs. 4325) when compared to the vanilla Transformer (\#1 vs. \#2). This revealed that PartialFormer has good practicability, and 2) In comparison to ODE Transformer, PartialFormer achieves similar inference speed and performance while significantly reducing memory consumption. This underscores PartialFormer's superiority over weight-sharing methods by effectively eliminating redundant computations.

\subsection{PG-FFNs vs. Vanilla Lightweight FFN}
In this section, we further emphasized PG-FFNs' superiority over vanilla lightweight FFN.
\paragraph{Settings.} We replaced the Transformer's FFNs with our PG-FFNs. In the decoder, we only integrated PG-FFNs for cross-attention, aligning with the vanilla Transformer. We set Transformer with reduced FFN hidden dimensions (384) as baseline.

\paragraph{Results.}
Table \ref{tab:PartialFormer_VARIANTS} (a) showcases the superior efficiency of our PG-FFNs. They outperform vanilla lightweight FFNs (26.82 vs. 26.07) with similar computational resources (40M vs. 41M, 7.7B vs. 7.7B). This is attributed to PG-FFNs' ability to maintain a large hidden dimension while using fewer parameters and computations, setting them apart from existing lightweight FFNs.

\begin{table*}[t!]
    \centering
    \renewcommand{\arraystretch}{1}
    \small

    \setlength{\tabcolsep}{1pt}
    \begin{minipage}[t]{0.6\linewidth}
        \centering
        \resizebox{0.9\linewidth}{!}{\begin{tabular}{lrrrrrrccc}
\toprule

\bf Model & $\bm{N}$-$\bm{M}$& $\bm{d}$ & $\bm{d_k}$  & $\bm{H}$& \bf MACs & \bf Param & \bf BLEU & \bf COMET \\
\midrule
Transformer	& 6-6 & 512 & 64	&8-8	&9.9B	&62M	&27.43 & 82.19	 \\
Transformer + LW FFNs	& 6-6 & 512 & 64	&8-8 &	7.7B	&41M	& 26.07 & 81.13
\\
Transformer + PG-FFNs	& 6-6 & 512 & 64	&8-8 &	7.7B	&40M	& 26.82 & 81.72	\\
\bottomrule
\end{tabular}}
        \caption*{(a) PG-FFNs vs. Vanilla FFNs.}
    \end{minipage}\hfill
    \begin{minipage}[t]{0.34\linewidth}
        \centering
        \setlength{\tabcolsep}{4pt}
        \resizebox{0.7\linewidth}{!}{\begin{tabular}{lcrc}
        \toprule
        \bf Model & \bf Param & \bf BLEU \\

        \midrule
        PartialFormer &68M & 29.56 \\
        PartialFormer-ODE &68M & 29.71 \\
        PartialFormer-GLU  &68M & 29.67 \\
        PartialFormer-DLCL  &68M & 29.88 \\
        \bottomrule
        \end{tabular}}
        \caption*{(b) Results of PartialFormer variants.}
    \end{minipage}
    \caption{(a) PG-FFNs offer a compelling alternative to
vanilla FFNs; (b) More results of PartialFormer variants. Metrics are reported on WMT’14 En-De.}
    \label{tab:PartialFormer_VARIANTS}
\end{table*}

\subsection{Combination with Existing Architectures }

We further investigated the adaptability and effectiveness of PartialFormer by applying it to three kinds of existing state-of-the-art architectures: 1) weight sharing methods~\cite{Lan2020ALBERT}, 2) gated linear units~\cite{Dauphin2017GLU} and 3) deep Transformer methods~\cite{wang-etal-2019-learning-deep}. We utilized the ODE Transformer~\cite{li-etal-2022-ode}, known for its parameter efficiency. Additionally, we selected Swi-GLU~\cite{shazeer2020glu} and DLCL~\cite{wang-etal-2019-learning-deep}.

Table \ref{tab:PartialFormer_VARIANTS} (b) shows the results. PartialFormer-DLCL achieved the highest performance, outperforming PartialFormer by 0.32 BLEU points, while PartialFormer-GLU showed the smallest improvement with an increase of 0.11 BLEU points. We attribute this to the fact that DLCL is an architecture-level modification addressing different issues from PartialFormer. In contrast, both GLU and ODE focus on the parameter-efficiency problem. Although ODE is also an architecture-level modification, its goal significantly overlaps with that of PartialFormer, leading to moderate performance improvements when combined. This indicates that PartialFormer already significantly enhances parameter efficiency, as adding ODE and GLU does not yield substantial performance gains.

\section{Related Work}
\paragraph{Lightweight Transformers}
Several strands of research have been dedicated to enhancing the parameter efficiency of the Transformer architecture, each taking a distinct approach to the problem at hand. The first category aims to mitigate redundancy directly through architectural innovations, employing more efficient transformation operations~\cite{Mehta2019DeFINEDF,Mehta2021Delight}, integrating disparate yet synergistic patterns~\cite{Wu2020Lite}, or leveraging neural architecture search techniques~\cite{So2019EvolvedTransformer}.
Another avenue of research explores weight sharing as a means of improving parameter efficiency, exemplified by the Universal Transformer's cross-layer parameter sharing strategy~\cite{Dehghani2019UniversalTransformer, reid-etal-2021-subformer-exploring}. Moreover, \citet{li-etal-2022-ode} introduced an ordinary differential equation-inspired weight-sharing approach to achieve higher performance. Different from them, our study focus on the design of lightweight FFN.

\paragraph{Multi-Branch Transformer}
The multi-branch strategy is widely used in Transformer design. Weighted Transformer~\cite{Ahmed2017WeightedTransformer} employs a multi-branch FFN, while Multi-attentive Transformer~\cite{Fan2020MultibranchAT}, Multi-units Transformer~\cite{yan-etal-2020-multi}, and Multi-Path Transformer~\cite{lin-etal-2022-multi-path,li-etal-2023-transformer} extend this concept to different components of the Transformer. Our PartialFormer can be viewed as a pure multi-branch architecture based on natural subspaces.

\paragraph{Scaling Strategy in Transformer}
Deepening~\cite{bapna-etal-2018-training, wang-etal-2019-learning-deep,li-etal-2020-shallow} and widening~\cite{Vaswani2017transformer, wu2021r} Transformer have been well-acknowledged as two strategies to improve the capacity of Transformer in literature. In this work, PartialFormer adopts two alternative strategies to improve the capacity: specifically, it enhances both the number of attention heads and the dimensions of each head.

\section{Conclusion}

In this paper, we present PartialFormer, a new parameter-efficient Transformer architecture that offers an alternative approach to the design of the lightweight FFN. By employing multiple small FFNs and leveraging matrix factorization techniques, PartialFormer effectively reduces the number of parameters in the FFN. Moreover, we propose two innovative operations to further efficiently enhance the model capabilities. Experimental results across various machine translation tasks showcase the significant performance improvements achieved by PartialFormer, while maintaining comparable parameter consumption.

\section*{Acknowledgments}
This work was supported in part by the National Science Foundation of China (No.62276056), the Natural Science Foundation of Liaoning Province of China (2022-KF-16-01), the Fundamental Research Funds for the Central Universities (Nos. N2216016 and N2316002), the Yunnan Fundamental Research Projects (No. 202401BC070021), and the Program of Introducing Talents of Discipline to Universities, Plan 111 (No.B16009).

\section*{Limitations}

Despite the potential advantages of Partialformer in terms of parameter utilization and performance within a limited parameter budget, it is important to note that the existing conclusions regarding its effectiveness have not been thoroughly examined in the context of large-scale datasets and a higher number of parameters. Further research is needed to validate the claims and assess the scalability of Partialformer in more challenging scenarios.

% \input{anthology.bbl}
% \input{acl.bbl}

% \bibliographystyle{acl_natbib}

\newpage

\appendix
\section{Detailed Setups of Experiments}
\label{sec:detailed_settup}

\subsection{Dataset}
Table \ref{tab:dataset_deatails} displays the statistics of all the 9 translation task.

\subsection{Training Details}
Table \ref{tab:training_ende_enfr_enro} and \ref{tab:training_wmt17} exhibits the training details on all translation tasks.

% The beam sizes were 4, 4, and 5 for En-De, En-Fr, and En-Ro tasks respectively. \textit{Length\_penalty} values of 0.6, 0.8, and 1.3 were applied to En-De, En-Fr, and En-Ro tasks respectively. For the WMT'17 benchmark, we set the beam size and \textit{Length\_penalty} to 4 and 1, respectively. We employed an ensemble of the last ten checkpoints.

% For the En-De task, we set the training updates to 70K with a batch size of 65536 under all model configurations. The warm-up updates are set to 16000 and the learning rate is set to 0.002. For the En-Fr task, the training updates are set to 100K with a batch size of 262144 under all model configurations. The warm-up updates and learning rate are set to 16000 and 0.002, 
% For the En-Ro task, the training updates are set to 25K with a batch size of 16384 under all model configurations. The warm-up updates and learning rate are set to 8000 and 0.002, respectively. Besides, we set the ratio of label smoothing to 0.1 for all tasks.
% For evaluation, We provide \textit{multi-BLEU} scores for all translation tasks. Additionally, we also provide COMET-22 scores for En-De and En-Fr tasks. During the evaluation, we set the \textit{beam} to 4, 4, and 5 in the En-De, En-Fr, and En-Ro tasks, respectively. As for the \textit{length\_penalty}, we set it to 0.6, 0.8, and 1.3 in the En-De, En-Fr, and En-Ro tasks, respectively.

\section{Implementation of Previous State-of-the-art Methods}
The accuracy of fairseq-based translation results can vary due to tokenization methods and other factors. To address fairness concerns, we re-implemented three state-of-the-art approaches in our codebase. To ensure absolute fairness, we employed the identical training strategy and data usage as in our PartialFormer model.

\paragraph{Data.} The dataset is sourced from Google's open release, featuring BPE operations totaling 32K. 
\paragraph{Training Strategy.} Our training strategy is the same as that of \citet{wang-etal-2019-learning-deep}, where 0.002 learning rate, 16000 warmup steps, pre-norm, relu\_dropout=0.1, attention dropout=0.1, 4096 tokens per GPUs (8 GPUs) and update the parameters every 2 steps.

\section{Ablation on Design of Decoder}
\label{sec_compare_decoder}

The design of the Decoder is a crucial component of the Transformer architecture due to its direct association with decoding. We evaluated three configurations: 1) Integrating PG-FFNs into both the decoder's self-attention and cross-attention, while halving the hidden dimension, 2) Incorporating PG-FFNs solely into the decoder's cross-attention, and 3) Incorporating PG-FFNs solely into the decoder's self-attention. 

Table \ref{tab:result_PG_FFNs_compare} exhibited the results on the WMT'14 En-De task. Our observations are as follows: 1) The first configuration yields the best performance, aligning with the insights from \citet{Gulati-etal-2020-conformer, Lu-etal-2020-Macaron}, 2) Using a single PG-FFN in each layer also delivers commendable results with a score of 29.21, and 3) Excluding PG-FFNs from the decoder's cross-attention results in erratic training, which is expected since there are no FFNs to handle the cross-attention features.

\section{Metric Definition}

\subsection{Measurement of Head Diversity}
Following \citet{li-etal-2018-multi-head}, we measure the head diversity as follows:
\begin{equation}
    D_\mathrm{output} = \exp(-\frac{1}{H^2}\sum_{i=1}^H\sum_{j=1}^H \frac{|O^i \cdot O^j|}{\Vert O^i \Vert \Vert O^j \Vert})
\end{equation}
    
During evaluation, we calculate the metric on all samples and average the values to obtain the final result.

\begin{table}[ht!]
\centering
\setlength{\tabcolsep}{1.5pt}
\resizebox{0.9\linewidth}{!}{\begin{tabular}{lrrrcc}
\toprule
% \specialrule{0em}{1.5pt}{0pt}
\multirow{2}{*}{\textbf{Dataset}}& \multicolumn{3}{c}{\textbf{Sentence}}& \multirow{2}{*}{\textbf{BPE}} & \multirow{2}{*}{\textbf{Vocab}}  \\
\cmidrule{2-4}
& \textbf{Train} & \textbf{Dev} & \textbf{Test} &  &  \\
\midrule
WMT'14 En-De & 4.5M&2999&3003 & 32K & 34040\\
WMT'14 En-Fr & 36M&26815&3003 & 32K & 37288\\
WMT'16 En-Ro  & 0.6M&1999& 1999 &  20K$\dag$&  19064\\
WMT'17 En-De & 5.9M&7998&3004 & 32K & 35488\\
WMT'17 De-En & 5.9M&7998&3004 & 32K & 35448\\
WMT'17 En-Fi  & 2.7M&4225& 3002 &  32K&  32584\\
WMT'17 Fi-En  & 2.7M&4225& 3002 &  32K&  32584\\
WMT'17 En-Lv  & 4.5M&2003& 2001 &  32K&  32368\\
WMT'17 Lv-En  & 4.5M&2003& 2001 &  32K&  32368\\
\bottomrule
\end{tabular}}
\caption{The details of datasets of 9 translation tasks.$\dag$: we follow the settings in \citet{li2022learning}.}
\label{tab:dataset_deatails}
\end{table}

\begin{table}[ht!]
\centering
\small
\setlength{\tabcolsep}{1pt}
\resizebox{\linewidth}{!}{\begin{tabular}{lccccc}
\toprule
& \bf En-De & \bf En-Ro & \bf En-Fr   \\
% {\bf Dataset} & \bf GPU$\times$Batch$\times$ UF & $\mathbf{LR}$ &{$\mathbf{Optim.}$} &\bf {$\mathrm{\mathbf{Adam}}_{\beta}$} &\bf Warmup&\bf Updates  \\
\midrule
GPUs & 8 & 4 & 8 \\
Batch Size & 4096 &4096 & 4096 \\
Update Frequency & 2 & 1 & 8 &\\
Optimer & Adam  & Adam  & Adam  \\
{$\mathrm{{Adam}}_{\beta}$}  & (0.9, 0.997)  & (0.9, 0.997)  & (0.9, 0.997)   \\
LR & 0.0020 &0.0020 & 0.0020 \\
LR scheduler  & inverse sqrt &inverse sqrt & inverse sqrt \\
Initial LR & 1$e^{-7}$ &1$e^{-7}$ & 1$e^{-7}$  \\
Total updates & 50K & 25K& 100K \\
Warmup updates & 16000 &8000 & 16000  \\
Weight decay & 0.0000 &0.0000 &  0.0000\\
Label smoothing & 0.1 & 0.1 & 0.1 \\
Dropout & 0.1  & 0.1  & 0.1  \\
Attention dropout & 0.1 & 0.1 & 0.1 \\
ReLU dropout & 0.1 & 0.1 & 0.1 \\
\toprule
\end{tabular}}
\caption{The training setups of WMT'14 En-De, WMT'16 En-Ro and WMT'14 En-Fr tasks.  }
\label{tab:training_ende_enfr_enro}
\end{table}

\begin{table}[ht!]
\centering
\small
\setlength{\tabcolsep}{1pt}
\resizebox{\linewidth}{!}{\begin{tabular}{lccccc}
\toprule
& \bf En-\{De, Lv\} & \bf \{De, Lv\}-En & \bf En-Fi  & \bf  Fi-En    \\
% {\bf Dataset} & \bf GPU$\times$Batch$\times$ UF & $\mathbf{LR}$ &{$\mathbf{Optim.}$} &\bf {$\mathrm{\mathbf{Adam}}_{\beta}$} &\bf Warmup&\bf Updates  \\
\midrule
GPUs & 8 & 8 & 8 & 8 \\
Batch Size & 4096 &4096 & 4096 & 4096 \\
Update Frequency & 2 & 1 & 1 & 4 \\
Optimer & Adam  & Adam  & Adam & Adam  \\
{$\mathrm{{Adam}}_{\beta}$}  & (0.9, 0.997)  & (0.9, 0.997)  & (0.9, 0.997) & (0.9, 0.997)   \\
LR & 0.0020 &0.0020 & 0.0020 & 0.0020 \\
LR scheduler  & inverse sqrt &inverse sqrt & inverse sqrt & inverse sqrt \\
Initial LR & 1$e^{-7}$ &1$e^{-7}$ & 1$e^{-7}$ & 1$e^{-7}$  \\
Total updates & 50K/17K & 50K/17K& 40K & 10K\\
Warmup updates & 16000 &16000 & 16000 & 16000  \\
Weight decay & 0.0000 &0.0000 &  0.0000 &  0.0000\\
Label smoothing & 0.1 & 0.1 & 0.1 & 0.1 \\
Dropout & 0.1  & 0.1  & 0.1  & 0.1  \\
Attention dropout & 0.1 & 0.1 & 0.1 & 0.1 \\
ReLU dropout & 0.1 & 0.1 & 0.1 & 0.1 \\
\toprule
\end{tabular}}
\caption{The training setups of WMT'17 benchmark.  }
\label{tab:training_wmt17}
\end{table}

\begin{table}[t!]
    \centering
    \renewcommand{\arraystretch}{1}
\centering
\small

\setlength{\tabcolsep}{1pt}
\begin{tabular}{lrrr}
\toprule
\bf Model  & \bf Param & \bf BLEU  \\
\midrule
PartialFormer		&68M	&29.56	 \\
\quad -PGFFNs in decoder self-AFFN		 &66M	& 29.21
\\
\quad -PGFFNs in decoder cross-AFFN		 &66M	& Failed \\
\bottomrule
\end{tabular}
    \caption{\textbf{Utilizing PG-FFNs in both the decoder's self-attention and cross-attention mechanisms is a preferable option.} BLEU points are reported in WMT'14 En-De task. }
    \label{tab:result_PG_FFNs_compare}
\end{table}

\begin{table}[t!]
\renewcommand{\arraystretch}{1}
\centering
\small
\setlength{\tabcolsep}{1.5pt}

\resizebox{0.9\linewidth}{!}{\begin{tabular}{lrc}
\toprule

\textbf{Model} 
& \bf Param  & \bf BLEU\\
\midrule
\textsc{Delight}~\cite{Mehta2021Delight} & 23M &  26.70\\
EdgeFormer~\cite{ge-etal-2022-edgeformer} & - &  26.90\\
Lite Transformer~\cite{Wu2020Lite} & - &  26.50\\
PartialFormer & 27M & \bf 27.50\\
\midrule
Evolved Transformer~\cite{So2019EvolvedTransformer} & 48M & 27.70 \\
\textsc{Delight}~\cite{Mehta2021Delight} & 37M &  27.60\\
ODE Transformer~\cite{li-etal-2022-ode} & 37M &  28.24\\
PartialFormer & 36M & \bf 28.35\\

\bottomrule
\end{tabular}}

    \caption{Comparison with state-of-the-art models of smaller capacities on the En-De task. }
    \label{tab:result_comparision_small}
\end{table}

\begin{table}[ht!]
    \centering
    \renewcommand{\arraystretch}{1}
\centering
\small
\setlength{\tabcolsep}{3pt}

\begin{tabular}{cccrcr}
\toprule
 \bf $\bm{A_G}$ & \bf $\bm{A_L}$ & \bf Param & \bf BLEU \\

% \multicolumn{5}{c}{\bf \textit{Previous NMT System}} \\
\midrule
RPR  &MHSA &62M & 35.76  \\
\bottomrule
\end{tabular}
\caption{Results of several PartialFormer variants on the En-De task.}
\label{tab:PartialFormer_A_G_VARIANTS}
\end{table}

\section{More Comparison with Previous Lightweight Transformer}

Table \ref{tab:result_comparision_small} presents a comprehensive comparison of previous lightweight Transformer models on the En-De task's test set, with a specific focus on operating within a smaller parameter budget. The results prominently showcase the outstanding performance of PartialFormer, even when faced with constraints on model capacity. This outcome further emphasizes the superior capabilities of PartialFormer in scenarios with limited resources.

\section{PartialFormer with Different $A_G$ for Small Dataset}
\label{sec:partial_different_A_G}
Table \ref{tab:PartialFormer_A_G_VARIANTS} showcases the results of PartialFormer on the WMT'16 En-Ro task, a small-scale translation dataset, specifically when $A_G$ is calculated using local attention~\cite{shaw-etal-2018-self}. Notably, these results reveal that by adopting such an approach, PartialFormer achieves an impressive BLEU score of 35.76. We hope this can shed lights on the area of model integration.

\section{Analysis on Token Uniformity}
\label{sec:analysis_tu}
Following \cite{Dong2021PureAttention, Wang2022AntiOversmoothing}, we measure the token uniformity among token representations. We use pearson correlation to compute it.

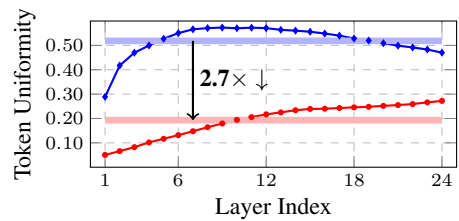
\begin{figure}[ht!]
    \centering
  \begin{tikzpicture}[]
  \pgfplotsset{set layers}
     \scriptsize{

    \begin{axis}[
	 at={(0.66\textwidth,0)},
      ymajorgrids,
      xmajorgrids,
      grid style=dashed,
      width=0.40\textwidth,
      height=.22\textwidth,
      legend style={at={(0.23,0.08)}, anchor=south west},
      xlabel={\small \makecell{Layer Index
      \\ }},
      ylabel={\small{Token Uniformity}},
      ylabel style={yshift=-1.5em},xlabel style={yshift=1.0em},
      yticklabel style={/pgf/number format/precision=2,/pgf/number format/fixed zerofill},
      ymin=0,ymax=0.6, ytick={ 0.1, 0.2, 0.3, 0.4, 0.5},
      xmin=0,xmax=25,xtick={1, 6, 12, 18, 24},
      legend style={yshift=-0.5em,xshift=4em,inner sep=1pt,legend plot pos=right,font={\footnotesize},cells={anchor=west}}
      ]
      % \draw[|-|,line width=0.6pt, black!80, dashed, thick] (62,31.23) -- (110, 31.23);
      % using "mark options" do more changes for marks

      \addplot[blue,mark=diamond*,mark size=0.5pt,thick,mark options={fill=blue,draw=blue,line width=1.25pt}] coordinates { (1, 0.28883352875709534) (2, 0.41732001304626465) (3, 0.4699432849884033) (4, 0.4995066225528717) (5, 0.5271627306938171) (6, 0.5504211783409119) (7, 0.5659817457199097) (8, 0.5709900259971619) (9, 0.5731505751609802) (10, 0.570284366607666) (11, 0.5726948976516724) (12, 0.5708319544792175) (13, 0.563860297203064) (14, 0.5604643225669861) (15, 0.5560687780380249) (16, 0.5486728549003601) (17, 0.5399861335754395) (18, 0.5283432602882385) (19, 0.5198920965194702) (20, 0.5089938640594482) (21, 0.498872846364975) (22, 0.49158498644828796) (23, 0.48306941986083984) (24, 0.47021931409835815)

      };
      % \addlegendentry{\scalebox{.6}{Transformer}}

      \addplot[blue!30,line width=2.5pt] coordinates { (1, 0.5186) (24, 0.5186)

      };

      \addplot[red, mark=otimes*,mark size=0.5pt,thick,mark options={fill=red,draw=red,line width=1.25pt}] coordinates { (1, 0.05046646296977997) (2, 0.06599926948547363) (3, 0.08231858164072037) (4, 0.10168235003948212) (5, 0.11680775880813599) (6, 0.1316044181585312) (7, 0.14767728745937347) (8, 0.16397510468959808) (9, 0.17946700751781464) (10, 0.19437670707702637) (11, 0.2057010680437088) (12, 0.21689032018184662) (13, 0.2251075953245163) (14, 0.23383928835391998) (15, 0.2391471564769745) (16, 0.24002841114997864) (17, 0.24297945201396942) (18, 0.24586531519889832) (19, 0.2480057328939438) (20, 0.2517806589603424) (21, 0.2551610767841339) (22, 0.2591455578804016) (23, 0.26618409156799316) (24, 0.2723335921764374)

      };

      \addplot[red!30, line width=2.5pt] coordinates { (1, 0.1932)  (24, 0.1932)

      };

      \draw[<-, thick] (axis cs:7, 0.1932) -- (axis cs:7,0.5186) ;
% \draw[->, thick] (axis cs:0, 0) -- (axis cs:5.5, 12.65) ;
\node[anchor=west,font=\footnotesize] at (axis cs:7, 0.3559) {\bf 2.7$\mathbf{\times \downarrow}$ };

      \end{axis}
      
     }

  \end{tikzpicture}
\vskip -0.1in
    \caption{Comparison of token uniformity (lower is better) in Transformer and PartialFormer. }
    \label{fig:analysis_tu}
\end{figure}

From Figure \ref{fig:analysis_tu}, we can observe that PartialFormer owns a lower token uniformity among token representations than the vanilla Transformer, revealing that PartialFormer can benefit from depth scaling efficiently~\cite{Dong2021PureAttention, Wang2022AntiOversmoothing}.

\definecolor{tiffanyblue}{RGB}{129,216,208}
\definecolor{bangdiblue}{RGB}{0,149,182}
\definecolor{kleinblue}{RGB}{0,47,167}
\definecolor{kabuliblue}{RGB}{26,85,153}
\definecolor{purple}{RGB}{138,43,226}

\section{Preliminary Experiments on Language Modeling}
We also evaluate the effectiveness of PartialFormer on the language modeling task.
\paragraph{Dataset.} For the language modeling task, we utilized the WikiText-103 dataset for evaluation. The training set comprises 103 million words from 28,000 articles, while the validation and test sets contain 218,000 and 246,000 words, respectively. We followed the data acquisition and preprocessing instructions from Fairseq~\cite{ott-etal-2019-fairseq}. 

\paragraph{Training \& Evaluation.} The training and evaluation settings adhere to the standard guidelines for language modeling in PyTorch~\cite{ott-etal-2019-fairseq}. We trained all models over 286,000 updates.

\paragraph{Results.}

Table \ref{tab:LM} exhibited results on the WikiText-103 task. PartialFormer surpasses the Adaptive Input model~\cite{Baevski2019AdaptiveInput} with a lower test perplexity of 19.87 compared to 21.11. Remarkably, PartialFormer achieves this with slightly fewer parameters (143M vs. 147M), demonstrating its efficiency and effectiveness as a language model for WikiText-103. We will present more comprehensive experiments in the future.

\begin{table}[t!]
\renewcommand{\arraystretch}{1}
\centering
\small
\setlength{\tabcolsep}{1.5pt}
\resizebox{\linewidth}{!}{
\begin{tabular}{lcccccc}
\toprule
\textbf{Model}  & $\bm{N}$ & $\bm{d}$& $\bm{d_k}$& $\bm{H}$& \bf Param & \bf Test PPL    \\
\midrule
Adaptive Input & 8 & 1024&128 & 8 & 147M & 21.11 \\
PartialFormer & 16 & 1024 & 256 & 4&  143M & \bf 19.87 \\
\bottomrule
\end{tabular}
}
\caption{\label{tab:LM}
Results on the WikiText-103 dataset.
}
\end{table}

\end{document}